\documentclass[10pt,twocolumn,letterpaper]{article}

\usepackage{iccv}
\usepackage{times}
\usepackage{epsfig}
\usepackage{graphicx}
\usepackage{amsmath}
\usepackage{amssymb}

\usepackage{booktabs}
\usepackage{multirow}
\usepackage{bbding}
\usepackage{pifont}
\usepackage{wasysym}
\usepackage{arydshln}
\usepackage{array}
\usepackage{subfigure}
\usepackage{colortbl} 
\usepackage{xcolor} 

\usepackage[pagebackref=true,breaklinks=true,letterpaper=true,colorlinks,bookmarks=false]{hyperref}

\usepackage[ruled,linesnumbered]{algorithm2e}
\usepackage{bm} 

\usepackage[capitalize]{cleveref}

\iccvfinalcopy

\def\iccvPaperID{****}

\ificcvfinal\fi

\begin{document}

\title{FSAR: Federated Skeleton-based Action Recognition with \\ Adaptive Topology Structure and Knowledge Distillation}

\author{
Jingwen Guo$^1$, ~
Hong Liu$^{1*}$, ~
Shitong Sun$^2$, ~
Tianyu Guo$^1$, ~
Min Zhang$^3$, ~
Chenyang Si$^{4*}$
\smallskip 
\\
$^1$Peking University, China ~
$^2$Queen Mary University of London, the UK \\
$^3$Harbin Institute of Technology, China ~
$^4$ Nanyang Technological University, Singapore
}

\maketitle

\ificcvfinal\thispagestyle{empty}\fi

\begin{abstract}
    Existing skeleton-based action recognition methods typically follow a centralized learning paradigm, which can pose privacy concerns when exposing human-related videos. 
    Federated Learning (FL) has attracted much attention due to its outstanding advantages in privacy-preserving. 
    However, directly applying FL approaches to skeleton videos suffers from unstable training. 
    In this paper, we investigate and discover that the heterogeneous human topology graph structure is the crucial factor hindering training stability. 
    To address this limitation, we pioneer a novel \textbf{F}ederated \textbf{S}keleton-based \textbf{A}ction \textbf{R}ecognition (\textbf{FSAR}) paradigm, which enables the construction of a globally generalized model without accessing local sensitive data. 
    Specifically, we introduce an Adaptive Topology Structure (\textbf{ATS}), separating generalization and personalization by learning a domain-invariant topology shared across clients and a domain-specific topology decoupled from global model aggregation.
    Furthermore, we explore Multi-grain Knowledge Distillation (\textbf{MKD}) to mitigate the discrepancy between clients and server caused by distinct updating patterns through aligning shallow block-wise motion features.  
    Extensive experiments on multiple datasets demonstrate that FSAR outperforms state-of-the-art FL-based methods while inherently protecting privacy.
\end{abstract}

\section{Introduction}
\label{sec:intro}

    \begin{figure}[t]
    \centering
       \includegraphics[width=1.0\linewidth]{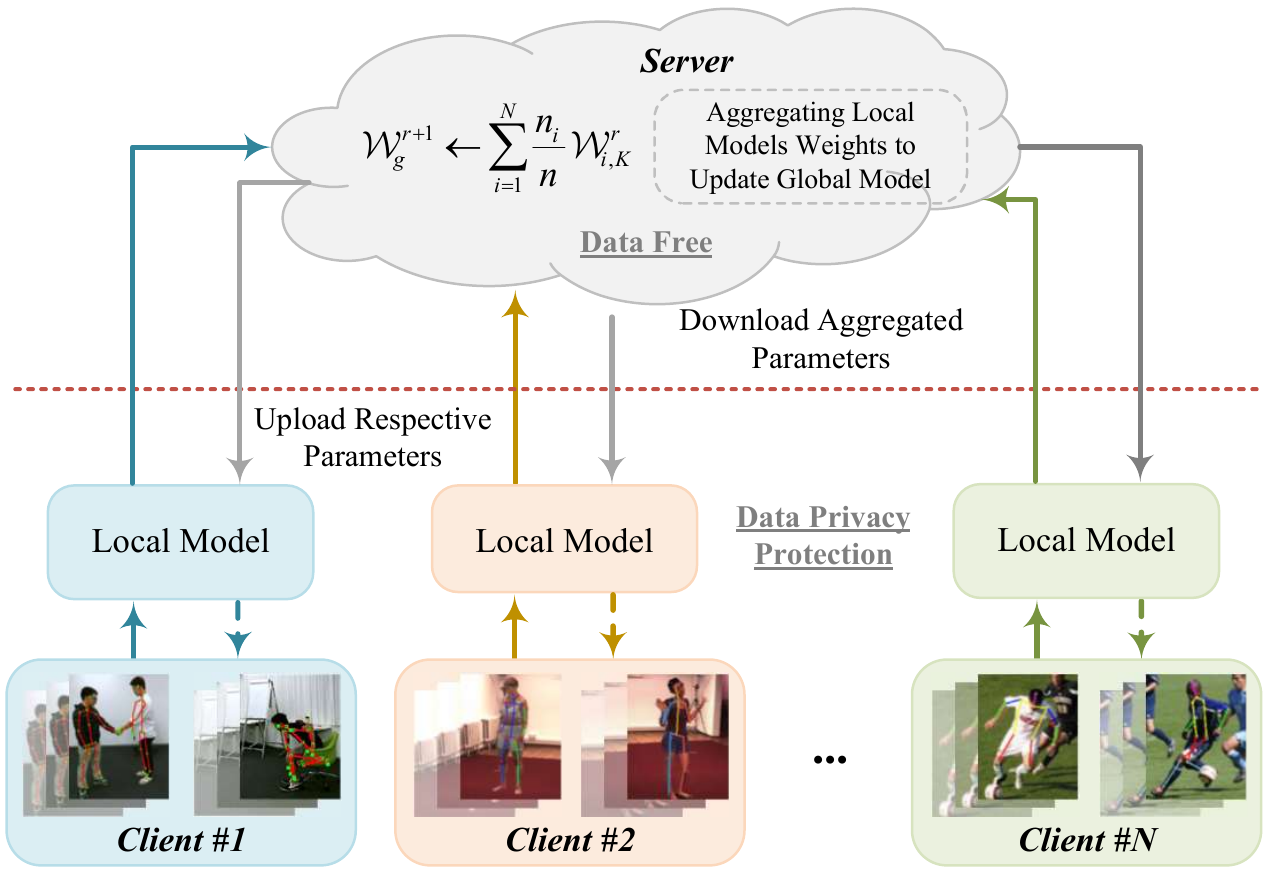}
       \caption{Illustration of Vanilla FSAR. 
       Each client optimizes a local model with non-shared and sensitive local skeleton videos, while the server obtain a global model by aggregating local model parameters without assessing any data. 
       The client-server collaborative learning is iterative to yield a global feature representation for out-of-the-box deployment with privacy protection.}
       \label{fig:motiva}
    \vspace{-13pt}
    \end{figure}
 
    Skeleton-based action recognition is a valuable research area with widespread applications in many fields~\cite{har1, har2, har3} such as human-robot interaction, intelligent security surveillance, and video understanding. Recent advancements in deep neural networks~\cite{dl-ske-st, dl-ske-st2} have shown significant progress in learning discriminative spatial and temporal features from skeleton sequences. 
    Though successful, these methods heavily rely on massively centralizing human skeleton videos, which directly and effectively pose privacy concerns due to the exposure of human-related information, like motion patterns, behavior tendencies, and personal identity. 
    For instance, Liao~\etal~\cite{person-identify,person-identify-2} have developed algorithms to identify individuals based on their body structures and walking styles.
    Therefore, centralized collection of such sensitive data exacerbates the risk of privacy disclosure for each local user site. 
    In response to the increasing awareness of personal data protection, decentralized training techniques are being developed, with federated learning being a powerful approach. 
    Thus in this work, we explore the application of federated learning to skeleton-based action recognition for privacy-preserving of skeleton data, which has rarely been investigated before.

    Federated Learning (FL) is a collaborative learning approach that aims to collectively learn from multiple decentralized edge devices or clients while preserving the security and privacy of local data~\cite{fedavg,fedbn}. 
    However, many effective FL techniques are image-based tasks, such as person re-identification~\cite{fed-reid}, medical image segmentation~\cite{fed-medical-rl}, and vision language navigation~\cite{fed-vln}. 
    To introduce FL schemes into the skeleton-based action recognition task, we follow the standard client-server architecture~\cite{fedavg, fl-kairouz} and construct a vanilla Federated Skeleton-based Action Recognition (Vanilla FSAR) paradigm. 
    As shown in~\cref{fig:motiva}, we consider multiple local clients having independent and non-overlapping category labels. 
    Each client is optimized with non-shared local data under the orchestration of a central server. The generalized server model is learned by aggregating the local model parameters without accessing sensitive local data. 
    Under this paradigm, vanilla FSAR optimizes a generic feature space from multiple and non-shared data silos while maintaining data privacy.

    \begin{figure}[t]
      \centering
        \subfigure[PKU I]{\includegraphics[width=4cm]{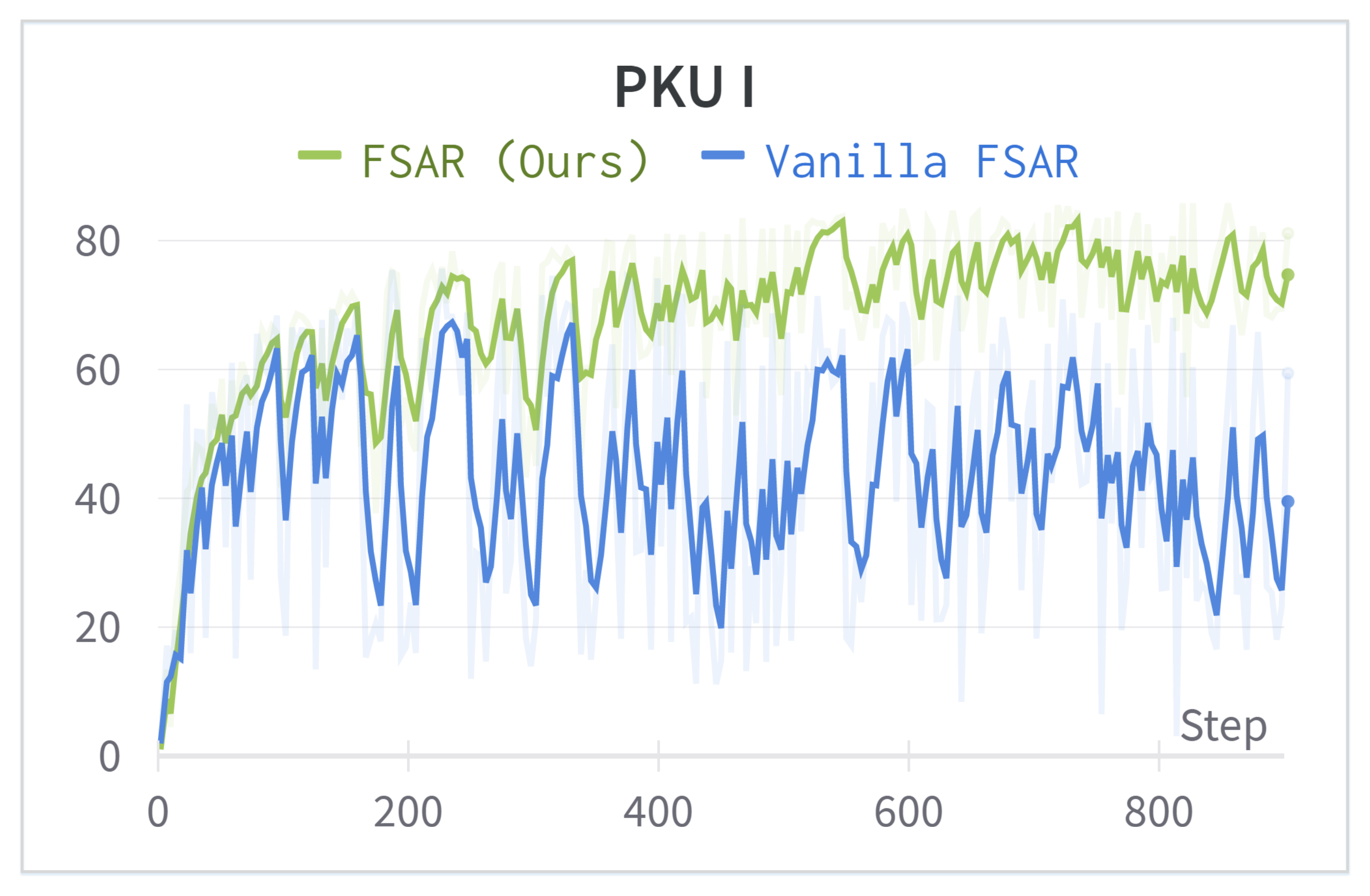}}
        \subfigure[NTU 60]{\includegraphics[width=4cm]{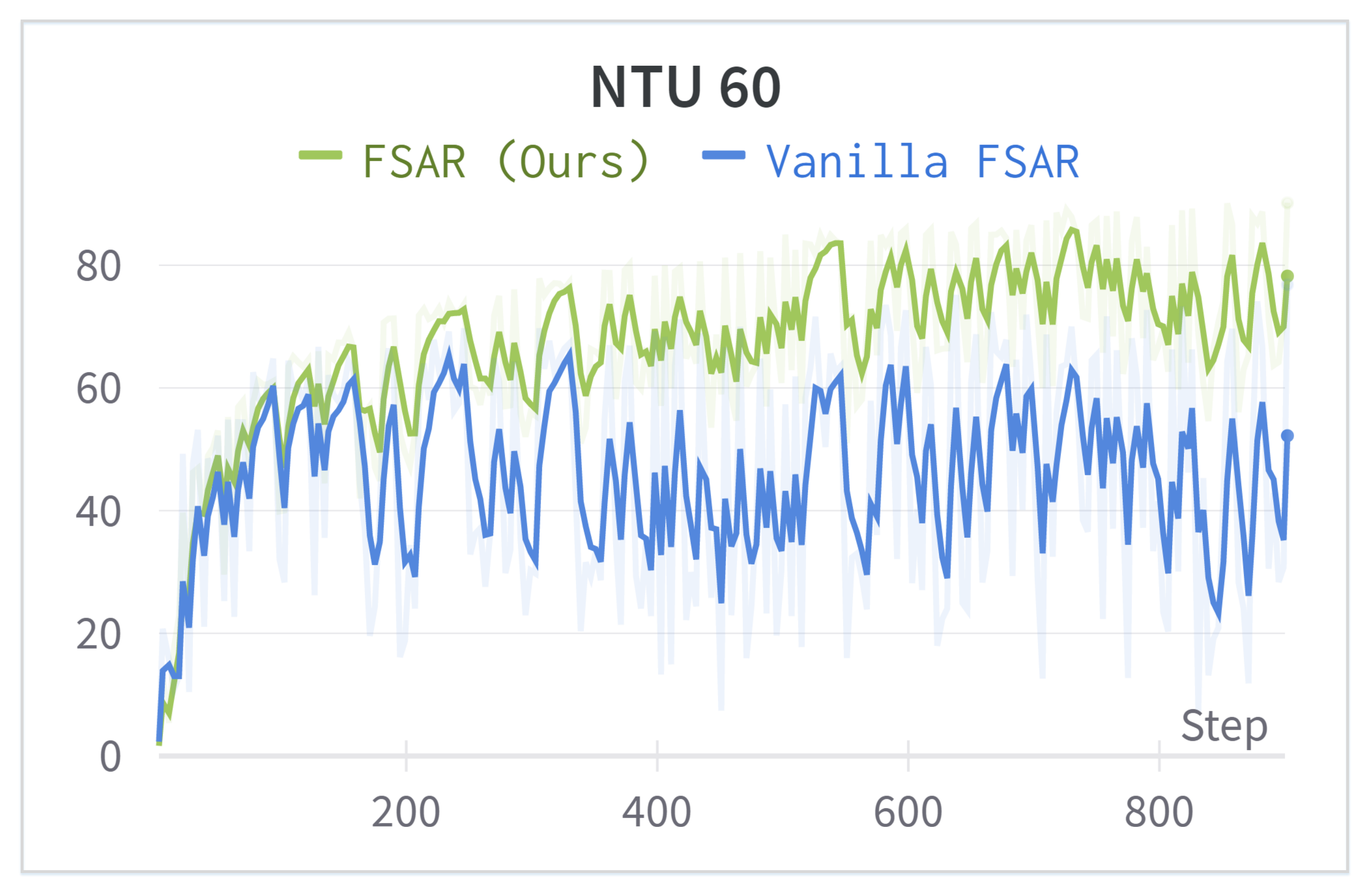}}
      \caption{Under the federated-by-dataset scenario, directly combining skeleton-based action recognition tasks with FL paradigms (blue ones) suffer from training instability due to the non-IID data distribution and heterogeneous human topology graph structures.}
      \label{fig:unstable}
      \vspace{-9pt}
    \end{figure}

    Due to the non-euclidean nature of skeleton data, Graph Convolution Networks (GCNs) have been effective models for skeleton-based tasks. One of the notable efforts is ST-GCN~\cite{stgcn} which models spatial temporal graph of skeleton sequences based on human topology structures. 
    Compared to CNN- or RNN-based methods, ST-GCN and its variants~\cite{gcn-ske-ar,gcn-ske-ar2,gcn-ske-ar3} achieve superior performance on skeleton-based action recognition. 
    Without loss of generality, we apply FL approaches to the ST-GCN model via the above client-server collaboration training paradigm as the straightforward  solution to address the privacy and security concerns. 
    Nevertheless, the direct combination can suffer from slow convergence and considerable fluctuations, as shown in ~\cref{fig:unstable}, which obstructs the model from generating feature representation suitable for efficient deployment with privacy protection. 
    Apart from the non-Independently Identically Distribution (non-IID)~\cite{noniid} of data from different clients, we identify the heterogeneous graph topologies across clients as a critical trigger for this phenomenon. 
    This structured element is dataset-specific, causing local models of different clients to gradually diverge from each other during local training. 
    It is known as the client drift problem in traditional FL~\cite{client-drift}, which can drastically damage the training performance of the global model when the data similarity decreases. 
    
    In light of this, we introduce FSAR, a novel benchmark for Federated Skeleton-based Action Recognition, to address the aforementioned client drift issues, especially for skeleton data.
    In FSAR, instead of finding one global model that fits the data distribution of all clients, an Adaptive Topology Structure (ATS) is proposed to inject modulated and customized elements into each client model. 
    The ATS learns the commonality of shared structure to \textit{improve the stability of FL training}, and preserves the unique structure of their own data to \textit{prevent current clients from being affected by other clients with different dataset scales}, respectively. 
    Moreover, we adopt learnable factors to automatically balance both topology structures on each client data since smaller datasets are more susceptible to large-scale datasets. 
    Apart from this, data heterogeneity across clients caused by various source domains also jeopardize the training stability and accuracy. 
    Generally, the features extracted by the shallower layer contain universal information (\eg, the connection between different joints), and the deeper features hold semantic information related to action labels (which is personalized and client-specific). 
    Therefore, a Multi-grain Knowledge Distillation (MKD) mechanism is further developed to reduce the feature variation of shallow layers, which decreases client divergences with respect to the server model and facilitates client-server communication on generalized messages.
    
    By leveraging both ATS and MKD, FSAR achieves remarkable performance and provides practical solutions for federated learning in skeleton-based action recognition.
    The contributions are summarized as follows:
    \vspace{-5pt}
    \begin{itemize}
    \setlength{\itemsep}{0pt}
    \setlength{\parsep}{0pt}
    \setlength{\parskip}{0pt}
        \item We take the lead in introducing federated learning into skeleton-based action recognition and present a novel benchmark FSAR to address privacy concerns in this field for the first time. 
        \vspace{2pt}
        \item We identify the heterogeneous graph topology structure as a major obstacle that causes training instability for skeleton data, and explore a novel Adaptive Topology Structure (ATS) to facilitate collaborative training between decentralized clients by learning domain-invariant and domain-specific topologies. 
        \vspace{2pt}
        \item The innovative Multi-grain Knowledge Distillation (MKD) mechanism is then explored by aligning shallow features to mitigate client-server divergence and further boost the accuracy.
        \vspace{2pt}
        \item Extensive experiments validate the effectiveness of FSAR and demonstrate that it achieves significant improvements over SOTA FL-based methods on several benchmark datasets with local data privacy protected.
    \end{itemize} 

    \begin{figure*}[ht]
    \centering
    \includegraphics[width=\linewidth]{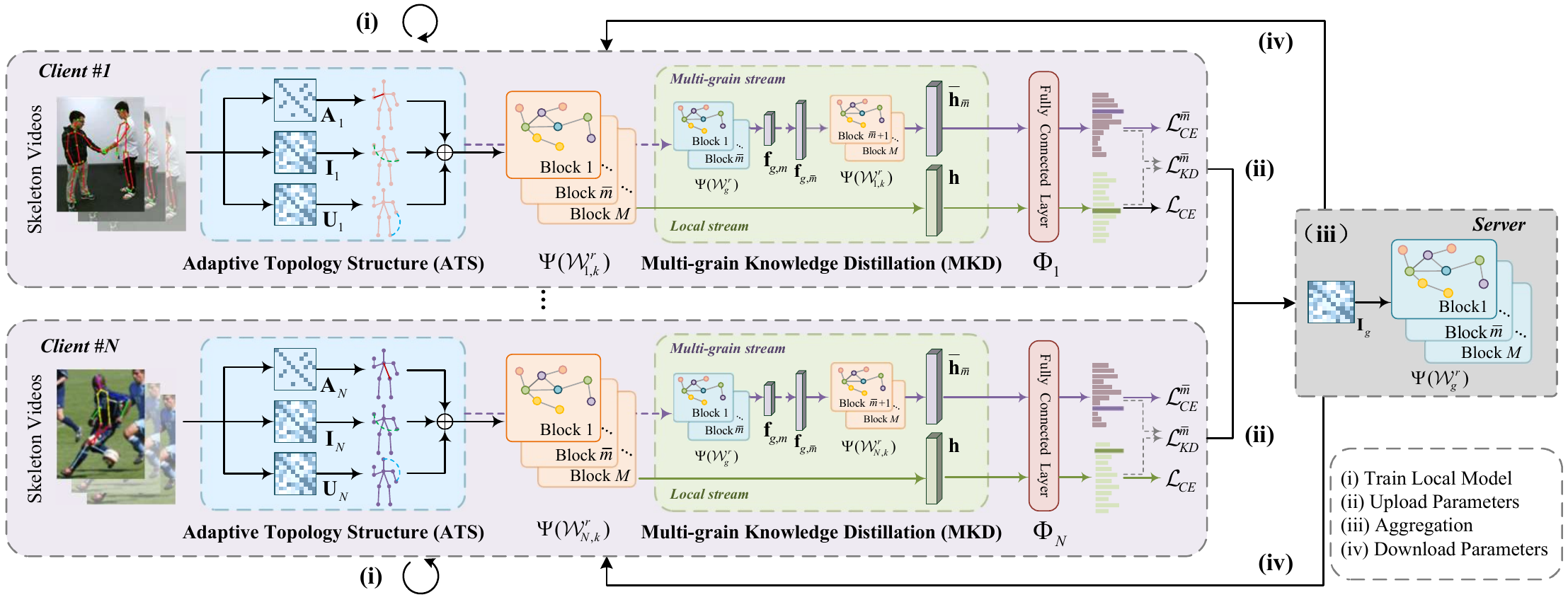}
    \caption{Overview of the proposed Federated Skeleton-based Action Recognition (FSAR). 
    The local clients are optimized with our proposed Adaptive Topology Structure (ATS) and Multi-grain Knowledge Distillation (MKD) modules on private data and then perform the client-server collaborative learning iteratively: 
    (i) clients train local models; (ii) clients upload parameters to server; (iii) server aggregates model parameters; (iv) clients download the aggregated models. 
    Moreover, the ATS module extracts the intrinsic structure information of heterogeneous skeleton data, and the MKD module bridges the divergence between the clients and the server.}
    \label{fig:illustration}
    \vspace{-9pt}
    \end{figure*}

\section{Related Work}  
\label{Related Work}

    \subsection{Skeleton-based Action Recognition}
    Skeleton-based Action Recognition aims to predict the action classes from skeleton sequences, whose joint coordinates are predicted from raw RGB video data by existing estimation algorithms. 
    Earlier approaches~\cite{manual1,manual2} focus on manually designing hand-crafted features and joint relationships, where the essential semantic connectivity of the human body is ignored. 
    Benefiting from the great representation ability of deep learning, some works rearrange the skeleton data into a grid-shaped structure and feed it directly into RNN~\cite{rnn, rnn-sichengyang} or CNN~\cite{cnn, cnn-liumengyuan} architectures. 
    Nowadays, inspired by human skeleton data being a natural topological graph, Graph Convolutional Networks (GCNs)~\cite{gcn, cz-aaai21-msgc, ggcn} have been adopted to exploit the underlying structure of joints. 
    ST-GCN~\cite{stgcn} is one of the most representative works, which introduces a spatio-temporal graph convolutional network to model the graph representation of each skeleton and capture temporal dynamics. 
    However, due to privacy concerns, contemporary centralized skeleton action recognition methods are significantly constrained by the limited scale of datasets available on the central server. 
    In this paper, we broaden the utility of these methods beyond centralized settings to improve quality and security.

    \subsection{Federated Learning Algorithms}
    Federated Learning is a decentralized learning framework that can train a global server model by aggregating the parameters of local clients~\cite{fedavg}. 
    Empirical works focus on better parameter aggregating strategies to alleviate feature drift caused by various domains and collaboratively transmit knowledge. 
    FedAVG~\cite{fedavg} proposes to aggregate the local model via a weight-based mechanism. 
    FedProx~\cite{fedprox} introduces a proximal term to restrict the local updates to be closer to the global model. 
    MOON~\cite{moon} utilizes the similarity between model representations to correct the local training. 
    FedAGM~\cite{fedagm} improves the stability and convergence of the server aggregation by sending the clients an accelerated model estimated with the global gradient. 
    FedEMA~\cite{fedema} updates local models adaptively using an exponential moving average update of the global model, where the decay rate is dynamically measured by model divergence. 
    Directly combining skeleton-based action recognition methods with the FL paradigms will encounter convergence problems due to the topological graph structures of the human body varying considerably among various source domains. 
    Existing FL optimization algorithms fail to effectively tackle the above issues, while our proposed FSAR shows superiority in this respect.

    \subsection{Action Recognition with Federated Learning}
    Federated Learning schemes~\cite{fl-kairouz,fedbn} have been explored in many computer vision tasks, such as medical image segmentation~\cite{fed-medical}, person re-identification~\cite{fed-reid-2}, and others. 
    When applying this paradigm to human action recognition tasks, 
    GraFehty~\cite{grafehty} constructs a similarity graph for each user to apply a GCN-based federated learning architecture to capture the inter-relatedness and closeness of human activities for classification tasks.  
    Mondal \etal~\cite{flhargcn} propose to transform time series sensor data into a graphical representation by taking non-overlapping time windows and aggregating the feature values of each time window. 
    Rehman \etal~\cite{fedvssl} and Dave \etal~\cite{spact} propose self-supervised learning frameworks under the FL paradigms for video action representation learning. 
    The above FL-based action recognition methods mainly focus on exploring the relationship between local data among clients or utilizing RGB data as input. 
    Unlike them, our method focuses on addressing the heterogeneity of the human skeleton topology graphs from different datasets domains.

\section{Methodology}
\label{Methodology}

    \subsection{Preliminary}
    In this work, we adopt the ST-GCN~\cite{stgcn} architecture as the local model since it performs well in capturing spatio-temporal relationships for skeleton videos. We first give a brief description of this basic backbone.
    
   \textbf{Spatial Graph.} 
    The input skeleton video $x \in \mathbb{R}^{T \times V}$ is first projected into hidden space, where $T$ and $V$ denote the number of frames per video and joints per person, respectively. 
    The human skeleton topology graph is first constructed as $\mathcal{G} = (\mathcal{V}, \mathbf{A})$, where $\mathcal{V} = \{v_i\}_{i=1}^V$ is the set of $V$ skeleton joints, and $\mathbf{A} \in \mathbb{R}^{V \times V}$ is the weighted adjacent matrix determining the connections between joints. 
    $A_{ij}=0$ if $v_i$ and $v_j$ are not connected, otherwise $A_{ij} \neq 0$.
    Followed by a spatial graph convolutional network, the input is operated as follows: 
     \begin{equation}
        \mathbf{f}_{out} = \sum_{s}^{S_v} {\mathbf{W}_s (\mathbf{f}_{in} A_s) \odot \mathbf{M}_s},
        \label{eq:gcn1}
    \end{equation}
    where $\mathbf{f}$ denotes the features, $S_v$ is the kernel size of the spatial dimension (empirically set as 3). 
    $\mathbf{M}_s$ is the attention map controlling the importance of each vertex, and $\mathbf{W}_s$ is the convolution parameters.
    
    \textbf{Temporal Filter.} After obtaining the structured information along the spatial dimension, a temporal convolutional filter is then employed to capture the motion patterns.

    \subsection{Vanilla FSAR} 
    \label{vanilla}
    
    \textbf{Overview.} We first establish the Vanilla Federated Skeleton-based Action Recognition (Vanilla FSAR) benchmark under the FL paradigms~\cite{fedavg}, as depicted in~\cref{fig:motiva}. 
    Suppose there are $N$ clients and one central server, where each client possesses its own private dataset and local model. 
    We optimize local client models with non-shared datasets and transfer their model parameters to the central server for aggregation. 
    The updated server model is then broadcast to each individual client for the next round. 
    This client-server collaborative learning is performed iteratively to learn a generalized central model.

    \textbf{Client updates.} For the $i$-th ($i \in N$) client, it is equipped with a backbone $\Psi(\mathcal{W}_{i,k}^{r})$ to project the input skeleton video into motion feature $\mathbf{h}_{i}$:
    \begin{equation}
            \mathbf{h}_i = \Psi(\mathcal{W}_{i,k}^{r};x),
        \end{equation} 
    where $\mathcal{W}_{i,k}^{r}$ is the parameters of backbone in the $k$-th local iteration of $r$-th global communication round. 
    Then the feature is fed into classifier $\Phi_i(\cdot)$ of the $i$-th client to produce the final output $\hat{y}$, \ie, $\hat{y} = \Phi_i(\mathbf{h}_i)$. 
    Following the iterative parameter updating mechanism in federated learning~\cite{fedavg},  
    the client model is optimized using SGD with a learning rate $\eta$. Hence, the $\mathcal{W}_{i,k}^{r}$ at the $k$-th iteration are updated:
    \begin{equation}
            \mathcal{W}_{i,k+1}^{r} \leftarrow \mathcal{W}_{i,k}^{r} - \eta \nabla \mathcal{W}_{i,k}^{r},
            \label{eq:client}
        \end{equation}
    where $\nabla \mathcal{W}_{i,k}^{r}$ is the set of gradient updates of the $i$-th client at the $k$-th local iteration of the $r$-th global round.
    
    \textbf{Server updates.} Through $K$ local iterations for updating client parameters, the server aggregates model parameters from clients to update the central backbone:
    \begin{equation}
            \mathcal{W}_{g}^{r+1} \leftarrow \sum_{i=1}^{N}{\frac{n_i}{n} \mathcal{W}_{i,K}^{r}},
            \label{eq:server}
    \end{equation}
    where $\mathcal{W}_g^{r}$ is the parameter of central backbone. 
    $\frac{n_i}{n}$ denotes the proportion of the data volume of the current client to the total data volume. 
    Then, the central backbone is used to reinitialize clients in the next round $\mathcal{W}_{i,0}^{r+1} \leftarrow \mathcal{W}_{g}^{r+1}$. 
    During inference, we combine the central backbone with the local classifier to perform action recognition.

    \begin{figure}[t]
        \centering
        \includegraphics[width=0.8\linewidth]{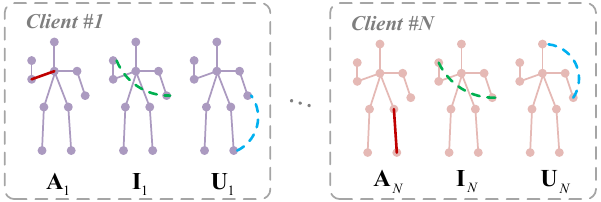}
        \caption{Illustration of ATS. $\mathbf{A}$ captures joint connection, $\mathbf{I}$ shares generalized structures, while $\mathbf{U}$ retains personalized knowledge.}
        \label{fig:aiu}
        \vspace{-8pt}
    \end{figure}
    
    \textbf{Limitations of Vanilla FSAR.} 
    As shown in blue curves in~\cref{fig:unstable} , Vanilla FSAR encounters training instability.
    Apart from the statistical heterogeneity caused by imbalanced distributions between clients, it still suffers from the following limitations: 
    (i) Insufficient knowledge mining for intrinsic human skeleton topology structures. (ii) Omission of the divergence between client and server. 
    Thus, we propose two key modules in~\cref{ATS} and ~\cref{MKD} to inject skeleton-linked-closely information into client-server communication under FL settings.

    \subsection{Adaptive Topology Structure}
    \label{ATS}
    The most crucial element in combining skeleton-based tasks with FL paradigms is the aforementioned human topology graphs, where the heterogeneous structures of the skeleton are closely related to the dataset. 
    Therefore, we propose an Adaptive Topology Structure (ATS) module for modeling skeleton videos to facilitate the server model capture inductive biases inherent in the topology graph from non-IID dataset domains.

    Apart from the manually set and domain-specific adjacent matrix $\mathbf{A}$, we introduce two more matrices to formulate ternary $(\mathbf{A}, \mathbf{I}, \mathbf{U})$ to mine generalized and personalized human skeleton topology at the same time. 
    The projection function in~\cref{eq:gcn1} is then revised as:
    \begin{equation}
        \mathbf{f}_{out} = \sum_{s}^{S_v} {\mathbf{W}_s \mathbf{f}_{in} (\alpha\mathbf{A}_s + \beta\mathbf{I}_s + \gamma\mathbf{U}_s)},
        \label{eq:gcn3}
    \end{equation}
    where $\alpha$, $\beta$, and $\gamma$ are coefficients to determine the importance of each matrix. 
    
    \textbf{Inflected Matrix (IM).} $\mathbf{I} \in \mathbb{R}^{V \times V}$ is introduced to alleviate the large training fluctuations caused by data heterogeneity by facilitating each client learning various skeleton joint connections from other clients progressively. 
    IM has the same dimension as $\mathbf{A}$ but is embodied with trainable parameters. 
    For each client and the central server, IM is updated in accordance manner of ~\cref{eq:client} and ~\cref{eq:server} with client-server collaborative training:
    \begin{equation}
        \mathbf{I}_{i,k+1}^{r} \leftarrow \mathbf{I}_{i,k}^{r} - \eta \nabla \mathbf{I}_{i,k}^{r},
    \end{equation}
    \begin{equation}
        \mathbf{I}_{g}^{r+1} \leftarrow \sum_{i=1}^{N}{\frac{n_i}{n} \mathbf{I}_{i,K}^{r}}.
    \end{equation}

    \textbf{Unique Matrix (UM).} $\mathbf{U} \in \mathbb{R}^{V \times V}$ is further introduced to avoid the current client being influenced by other clients when their dataset scales vary significantly, by retaining personalized skeleton topology at the local. 
    Unlike IM, UM updates individually, without communication or aggregation at the server. Namely, the parameters of UM are excluded from the global server and kept in local clients only.

    Notice that our proposed IM and UM are explicitly designed for addressing the client drift issues in FL tasks, different from previous works~\cite{agcn, asgcn}, which use attention mechanisms to capture connections between joints for each individual dataset. 
    As shown in~\cref{fig:aiu}, IM participates in aggregation and is shared across clients to learn domain-invariant topology and transmit common skeleton connections. 
    Furthermore, UM is preserved locally to maintain client-specific topology and prevent current clients from being affected by other datasets largely. 

    \begin{figure}[t]
        \centering
        \includegraphics[width=1.0\linewidth]{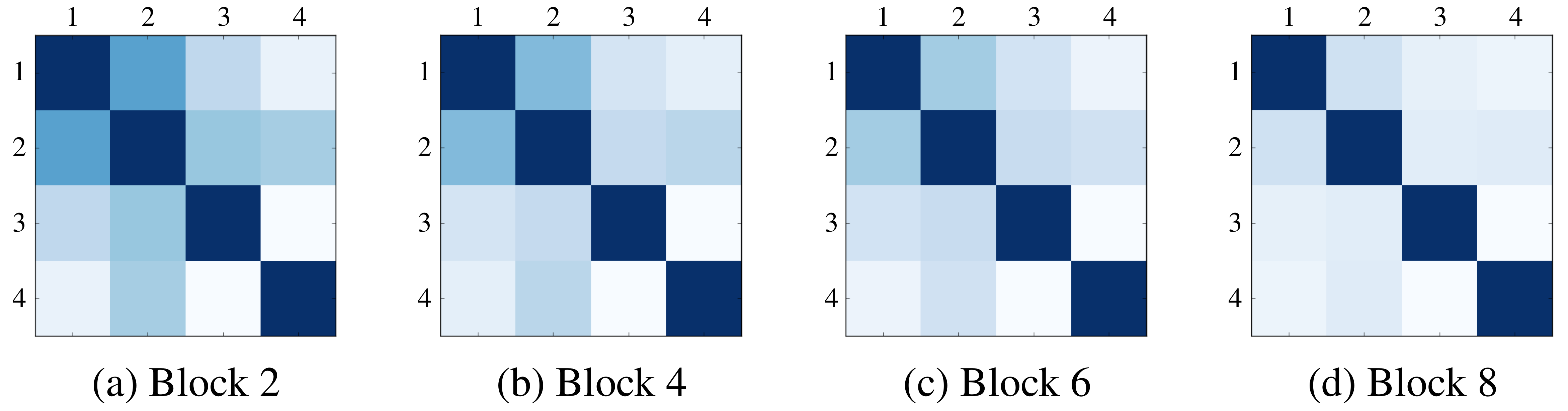}
        \caption{Visualization of the CKA similarities of different level ST-GCN blocks from clients (coordinate axes refer to the $No.$ of clients). 
        The deeper the block is, the fewer similarities they gain.}
        \label{fig:cka}
        \vspace{-8pt}
    \end{figure}

    \begin{table*}[t]
        \centering
        \caption{Comparison with different combinations of Vanilla FSAR and existing FL optimization methods, ablation studies for different proposed modules, and performance with different baseline models. * Vanilla FSAR utilizes the FedAVG~\cite{fedavg} for aggregation.}
        \resizebox{\linewidth}{!}{
            \begin{tabular}{clccccc}
            \toprule
              \multirow{2}{*}{Backbone} & \multirow{2}{*}{Model} & \multicolumn{5}{c}{Client (Linear-Accuracy, $\%$)} \\  \cmidrule{3-7}
              \multicolumn{1}{c}{}      &                        & PKU MMD I   & PKU MMD II   & NTU RGB+D 60  & NTU RGB+D 120 & UESTC   \\ \midrule
              \multirow{8}{*}{STGCN~\cite{stgcn}}    
                                        & Vanilla FSAR*     & 77.77   & 48.89   & 81.08   & 74.77   & 80.91        \\
                                        & FedProx~\cite{fedprox} & 76.85 \textcolor{blue}{(-0.92)}    & 47.81 \textcolor{blue}{(-1.08)}  & 81.37 \textcolor{red}{(+0.29)}  & 75.32 \textcolor{red}{(+0.55)}  & 80.88 \textcolor{blue}{(-0.03)}        \\
                                        & FedBN~\cite{fedbn}     & 78.13 \textcolor{red}{(+0.36)}    & 46.49 \textcolor{blue}{(-2.40)}  & 78.45 \textcolor{blue}{(-2.63)}   & 72.89 \textcolor{blue}{(-1.88)}   & 79.72 \textcolor{blue}{(-1.19)}       \\
                                        & FedAGM~\cite{fedagm}   & 78.05 \textcolor{red}{(+0.28)}    & 49.13 \textcolor{red}{(+0.24)}  & 81.56 \textcolor{red}{(+0.48)}   & 76.00 \textcolor{red}{(+1.23)}   & 79.27 \textcolor{blue}{(-1.64)}       \\
                                        & MOON~\cite{moon}       & 75.69 \textcolor{blue}{(-2.08)}    & 46.88 \textcolor{blue}{(-2.01)}  & 79.45 \textcolor{blue}{(-1.63)}   & 73.70 \textcolor{blue}{(-1.07)}   & 81.48 \textcolor{red}{(+0.57)}       \\ 
                                         & \cellcolor{gray!15}\textbf{FSAR (Ours)}   & \cellcolor{gray!15}\textbf{81.96 \textcolor{red}{(+4.19)}}    & \cellcolor{gray!15}\textbf{56.30 \textcolor{red}{(+7.41)}}  & \cellcolor{gray!15}\textbf{91.30 \textcolor{red}{(+10.22)}}   & \cellcolor{gray!15}\textbf{84.31 \textcolor{red}{(+9.54)}}   & \cellcolor{gray!15}\textbf{91.88 \textcolor{red}{(+10.97)}}   \\       \cdashline{2-7}
                                        & FSAR  w/o ATS          & 79.73 \textcolor{red}{(+1.96)}       & 52.38 \textcolor{red}{(+3.49)}       & 84.45 \textcolor{red}{(+3.37)}       & 76.83 \textcolor{red}{(+2.06)}       & 84.56 \textcolor{red}{(+3.65)}    \\
                                        & FSAR  w/o MKD          & 81.36 \textcolor{red}{(+3.59)}    & 53.83 \textcolor{red}{(+4.94)}     & 89.50 \textcolor{red}{(+8.42)}     & 83.34 \textcolor{red}{(+8.57)}      & 90.33 \textcolor{red}{(+9.42)}    \\  \midrule
              \multirow{2}{*}{CTR-GCN~\cite{ctrgcn}}  
                                        & Vanilla FSAR*          & 81.85    & 51.87     & 82.91    & 76.48    & 81.37   \\
                                        & \cellcolor{gray!15} FSAR (Ours)            & \cellcolor{gray!15} 84.19 \textcolor{red}{(+2.34)}     & \cellcolor{gray!15} 59.04 \textcolor{red}{(+7.17)}     & \cellcolor{gray!15} 91.64 \textcolor{red}{(+8.73)}     & \cellcolor{gray!15} 84.91 \textcolor{red}{(+8.43)}    & \cellcolor{gray!15} 92.60 \textcolor{red}{(+11.23)}   \\ \midrule
              \multirow{2}{*}{MST-GCN~\cite{cz-aaai21-msgc}}  
                                        & Vanilla FSAR*          & 79.01    & 50.14     & 81.98    & 74.93    & 80.89   \\
                                        & \cellcolor{gray!15} FSAR (Ours)            & \cellcolor{gray!15} 82.48 \textcolor{red}{(+3.47)}    & \cellcolor{gray!15} 57.45 \textcolor{red}{(+7.31)}     & \cellcolor{gray!15} 90.07 \textcolor{red}{(+8.09)}    & \cellcolor{gray!15} 83.53 \textcolor{red}{(+8.60)}    & \cellcolor{gray!15} 91.53 \textcolor{red}{(+10.64)}    \\ \midrule
              \multirow{2}{*}{MS-G3D~\cite{msg3d}}   
                                        & Vanilla FSAR*          & 78.17    & 51.89     & 80.15    & 75.02    & 81.26    \\
                                        & \cellcolor{gray!15}FSAR (Ours)            & \cellcolor{gray!15} 82.83 \textcolor{red}{(+4.66)}    & \cellcolor{gray!15} 57.98 \textcolor{red}{(+6.09)}     & \cellcolor{gray!15} 90.98 \textcolor{red}{(+10.83)}    & \cellcolor{gray!15} 84.84 \textcolor{red}{(+9.82)}    & \cellcolor{gray!15} 91.79 \textcolor{red}{(+10.53)}    \\ \bottomrule
                                           
            \end{tabular}
        }
        \label{tab:general_results}
        \end{table*}

    \subsection{Multi-grained Knowledge Distillation}
    \label{MKD}
    We visualize the Centered Kernel Alignment (CKA) similarities~\cite{cka} in ~\cref{fig:cka}, finding that the divergence of features between clients from deeper blocks gets much more considerable than shallower ones.  
    This enlightens us to propose Multi-grain Knowledge Distillation (MKD) mechanism by taking the shallow block-wise features from the central server model as teacher knowledge to supervise the local training of individual clients. 
    In this way, the shallow blocks of the client model get generalized while the deep retain personalized. 

\begin{figure}[t]
        \centering
        \includegraphics[width=1.0\linewidth]{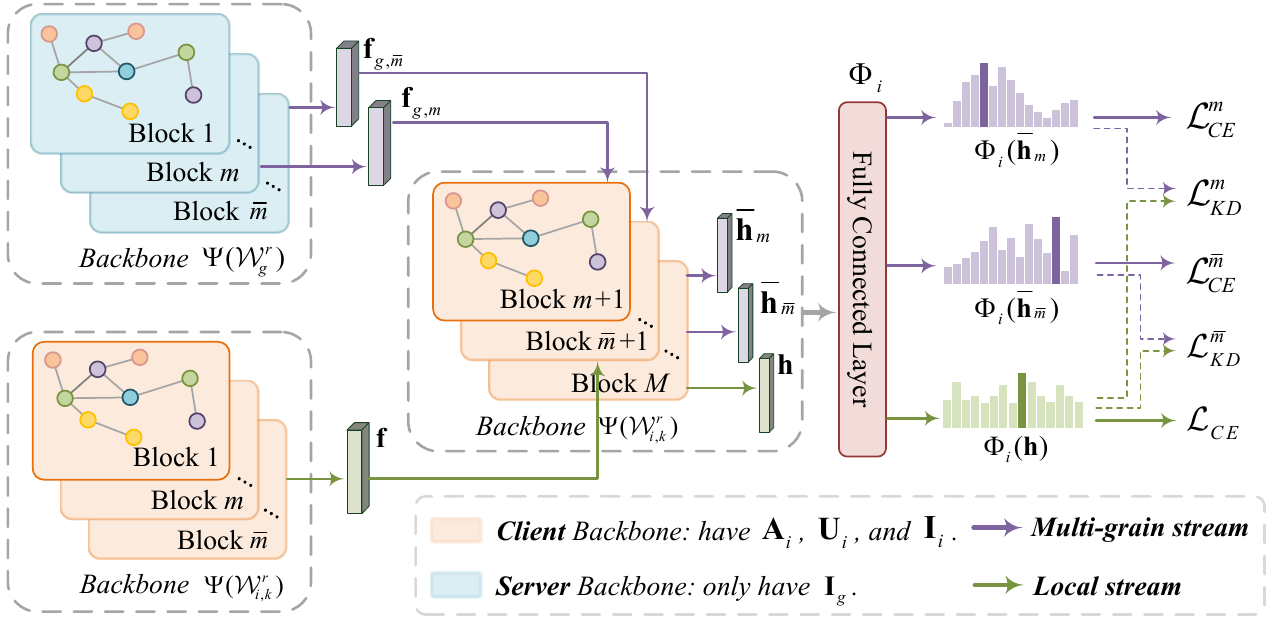}
        \caption{Illustration of MKD module. Both the client and server models are divided into $M$ exclusive blocks based on the depths and the motion feature sizes. The central server model is taken as the empirical teacher to supervise the local client training via shallow block-wise feature alignment.}
        \label{fig:mfkd}
        \vspace{-8pt}
    \end{figure}

    As shown in ~\cref{fig:mfkd}, the multi-grain stream is introduced apart from the original local stream. 
    Instead of directly applying feature-level supervision between the server and clients, MKD gradually adopts the shallow blocks of server model to replace the corresponding blocks of client models and feeds the shallow features from the server into the bottom block of the client. 
    Then, the predictions of this combined network are used to be teacher supervision signals. 
    Since the server model is shared for different clients, directly requiring similar outputs between clients and their corresponding teacher supervision predictions can align the shallow representations of client models.

    More specifically, given the input skeleton sequence $x$ and the $i$-th client model, the local stream extracts the motion features $\mathbf{h}$. 
    For the multi-grain stream, we use the first $m$ blocks of the server model to produce the middle-layer feature $\mathbf{f}_{g,m}$, which will be fed into the shared bottom block of the $i$-th client to obtain the teacher representations $\bar{\mathbf{h}}_m$. 
    Finally, $\mathbf{h}$ and $\bar{\mathbf{h}}_m$ are fed into the $i$-th client classifier to predict their probabilities. 
    To align the shallow representations of client models, we require each client to achieve a similar prediction with the teacher representations $\bar{\mathbf{h}}_m$. 
    Therefore, the multi-grain knowledge distillation loss is defined as follows for the client model additionally: 
    \begin{equation}
            \mathcal{L}_{KD} = \sum_{m=1}^{\bar{m}} {KL( \Phi_i (\bar{\mathbf{h}}_m), \Phi_i(\mathbf{h}))},
    \end{equation}
    where $\bar{m}$ denotes the granularity of the feature we want to align, $KL(\cdot)$ is Kullback Leibler divergence, and $\Phi_i(\cdot)$ is the  $i$-th client classifier. 
    
    Accordingly, the overall classification loss is formulated as a combination of losses in two streams as follows:
        \begin{equation}
            \mathcal{L}_{CE} = CE(\Phi_i(\mathbf{h}), y)
                               + \sum_{m=1}^{\bar{m}} {CE(\Phi_i (\bar{\mathbf{h}}_m), y)},
        \end{equation}
    where $y$ is truth-label, and $CE(\cdot)$ is Cross Entropy. 
    Note that the server model does not contain UM but only IM. 
    It makes sense since the IM is obligated to alleviate the divergence between client and server while the UM is endowed to retain personalized information locally.

    \subsection{Optimization Strategy} 
    During training, we use momentum averaging strategy~\cite{fedagm} to accelerate the model training. 
    Apart from the global gradient-guided updates, a regularization term is added for each client by calculating the $L2-norm$ between the parameters of the current client and central server:
    \begin{equation}
         \mathcal{L}_{Reg} = \frac{1}{2} || \mathcal{W}_{g}^{r} - \mathcal {W}_{i,k}^{r} || ^ 2.
    \end{equation}

    Therefore, in each client-server communication round, the individual $i$-th client is trained in an end-to-end manner using the combination of the above losses: 
    \begin{equation}
        \mathcal{L}(\mathcal{W}_i) = \lambda_{1} \mathcal{L}_{CE} + \lambda_{2} \mathcal{L}_{KD} + \lambda_{3} \mathcal{L}_{Reg},
        \label{eq:loss}
    \end{equation}
    where $\lambda_1$, $\lambda_2$ and $\lambda_3$ are the coefficients to balance the loss. 

    \begin{figure*}[t]
        \centering
        \subfigure[PKU II]{\includegraphics[width=5.4cm]{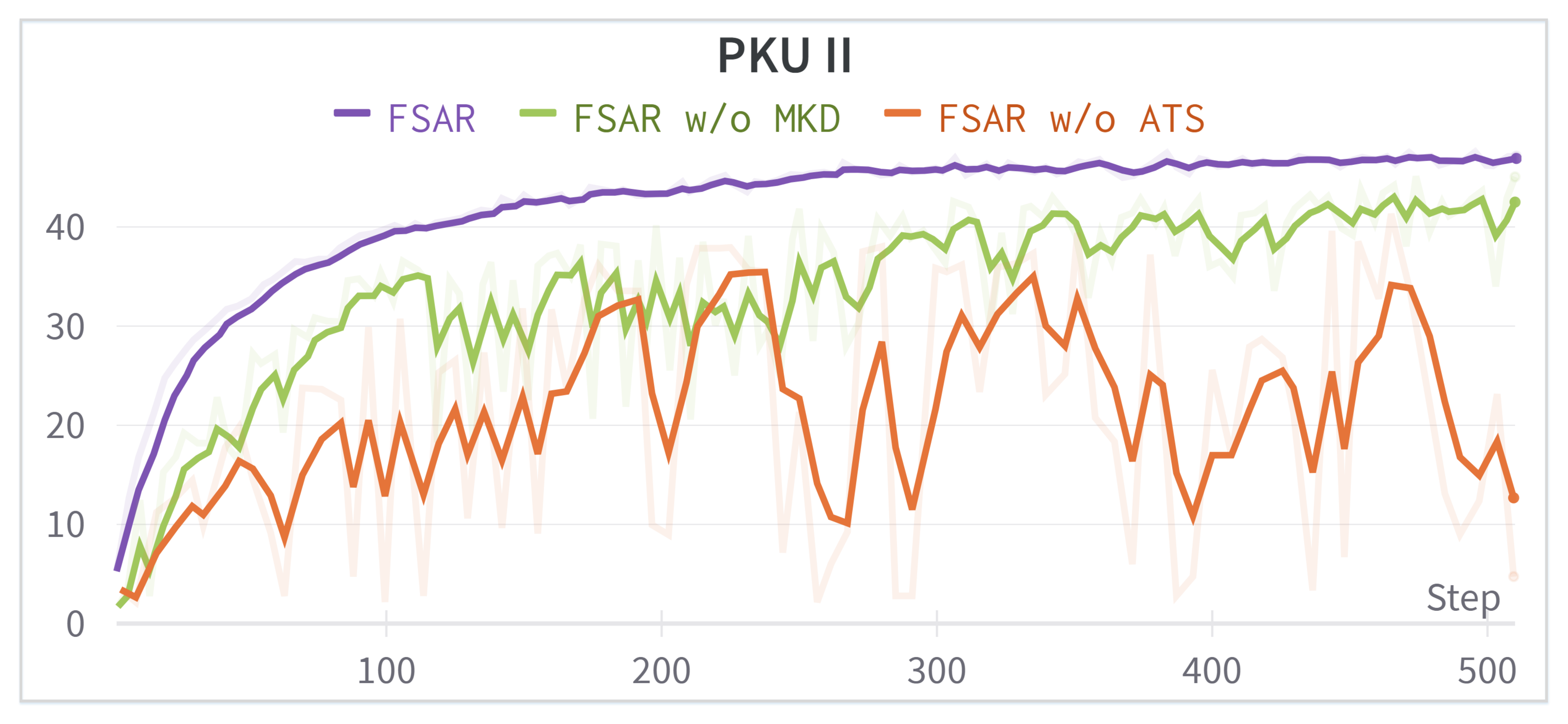}}      \hspace{2pt}
        \subfigure[NTU 120]{\includegraphics[width=5.4cm]{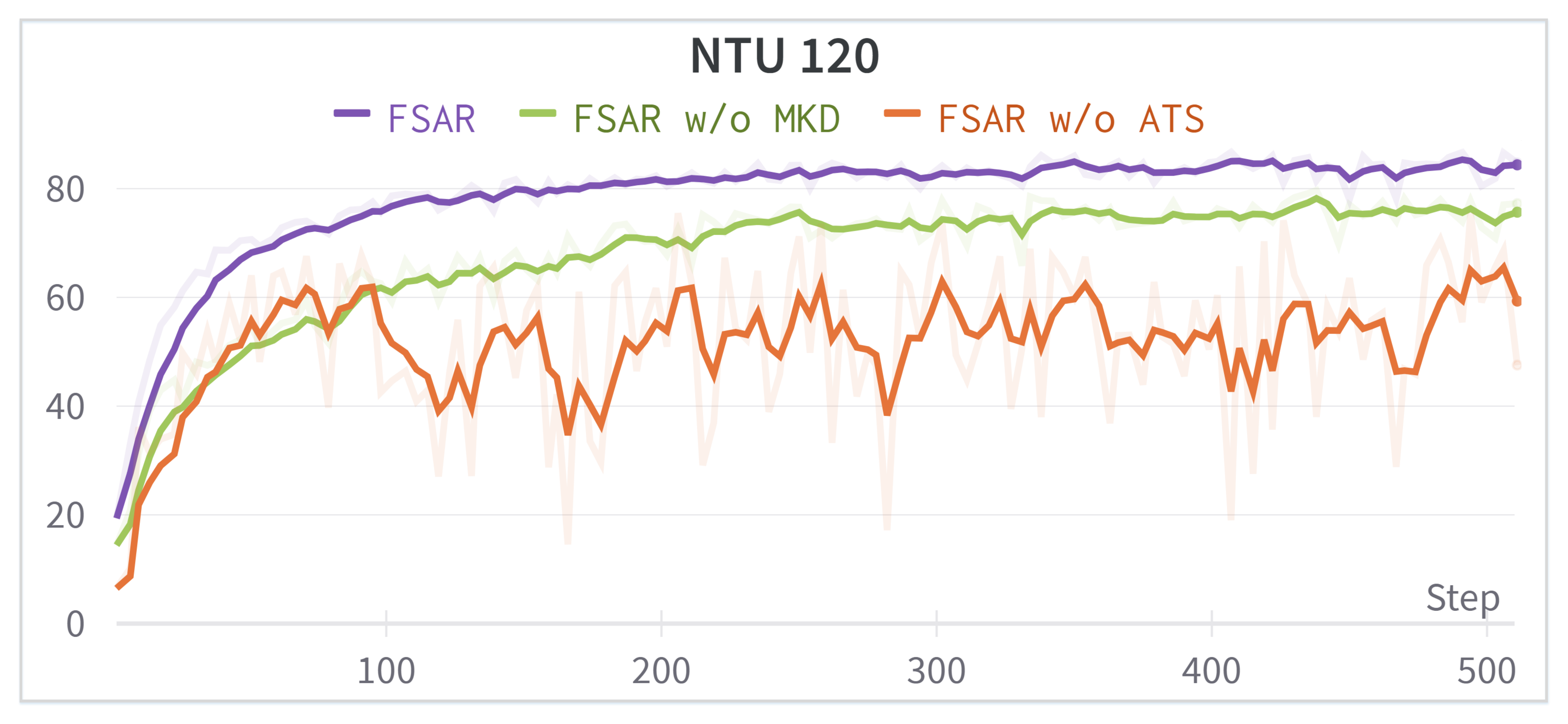}}   \hspace{2pt}
        \subfigure[UESTC]{\includegraphics[width=5.4cm]{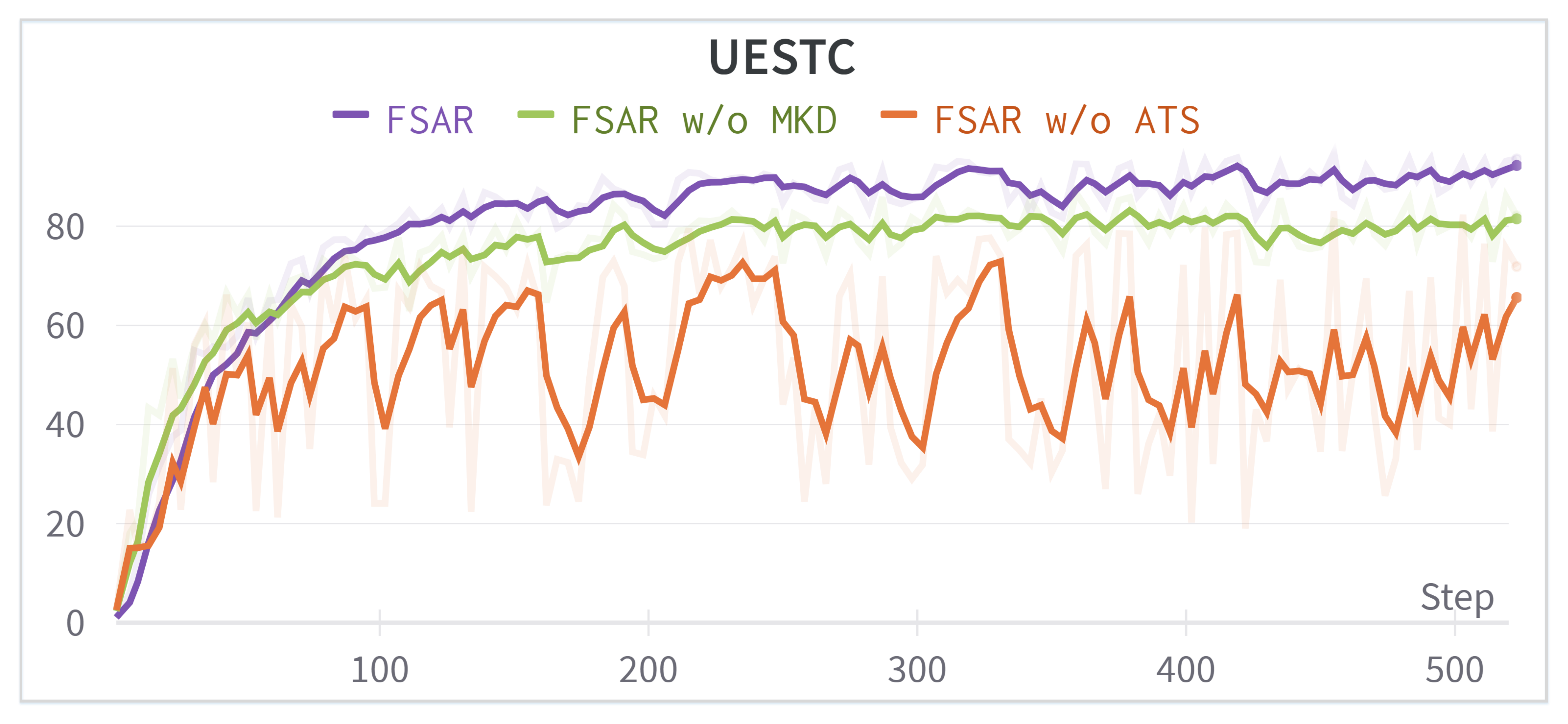}}

    \caption{Comparison of Linear-Accuracy test curves between FSAR (purple), FSAR w/o MKD (green), and FSAR w/o ATS (orange). The results are reported on PKU II (left), NTU 120 (middle), and UESTC (rigth) datasets. Our proposed FSAR achieves faster convergence and higher accuracy on all datasets.}
    \label{fig:test_accuracy}
    \vspace{-8pt}
    \end{figure*}

\section{Experiments}

    \subsection{Datasets and Federated Scenarios}  
    We evaluate our method on widely-used datasets of various scales for skeleton-based action recognition:

    \textbf{NTU RGB+D} ~\cite{nturgbd,nturgbd120} is a large-scale human action recognition dataset. 
    The NTU 60 contains 56,880 skeleton action sequences, performed by 40 volunteers and categorized into 60 classes. 
    The NTU 120 extends NTU 60 with additional 57,367 skeleton sequences over 60 extra action classes. 
    In total 113,945 samples over 120 classes are performed by 106 volunteers.
    
    \textbf{PKU MMD} ~\cite{pkummd} contains 20,000 action sequences covering 51 action classes. 
    It consists of two subsets. 
    PKU I is an easier version for action recognition, while PKU II is more challenging with more noise caused by view variation.
    
    \textbf{UESTC} ~\cite{uestc} is a newly built dataset for arbitrary-view action analysis. 
    There are a total of 25,600 sequences of 40 action classes performed by 118 subjects.

    We conduct our experiments under the standard \textit{federated-by-dataset} scenario, where each client constructs its own private dataset and then collaboratively conducts federated learning with the central server. 
    This conforms to the definition of non-IID data distribution properties. During inference, we combine the central server backbone with local client classifiers to perform action recognition. 
    Furthermore, we employed \textit{adaptated-to-unseen-dataset} scenario to analyze the generalization of our FSAR. 
    Specifically, we randomly select four of the total five datasets for training the global model and then test the server backbone on the remaining one unseen dataset. 
    We use Linear-Accuracy and KNN-Accuaracy evaluation protocols for the above two scenarios, respectively.

    \subsection{Implementation Details}
    All the experiments are conducted in the PyTorch~\cite{pytorch} framework. 
    For data pre-processing, we resize each skeleton sequence to the length of 50 frames and adopt the code of Yan~\etal~\cite{stgcn} for data augmentations. 
    Cross-subject evaluation protocol is adopted for NTU and PKU datasets. 
    For the backbone network ST-GCN, the batch size is 128, the feature dimension is 128, and the channel is 1/4 of the original setting. 
    For optimization, we use SGD with momentum (0.9) and weight decay (0.0001) for local clients training. 
    The overall FSAR model is trained with the local epoch $K=1$ and the total communication rounds $R=300$. 
    The loss weights are set as $\lambda_1 = \lambda_2 = 1$ and $\lambda_3=0.1$.

    \begin{table}[t]
        \centering
        \caption{KNN-Accuracy (\%) on unseen datasets. Unseen dataset refers to the individual one dataset used for testing, where the remaining four datasets are used for training the global model.}
        \resizebox{\linewidth}{!}{
            \begin{tabular}{lccccc}
            \toprule
                \multirow{2}{*}{Settings} & \multicolumn{5}{c}{Unseen datasets (KNN-Accuracy, \%)} \\   \cmidrule{2-6}
                                          &  PKU I     & PKU II    & NTU 60    & NTU 120    & UESTC \\ \midrule
                Vanilla FSAR              & 75.81      & 45.18     & 81.62     & 71.51      & 84.89   \\
                \rowcolor{gray!15} \textbf{FSAR (Ours)}      & \textbf{78.39}     & \textbf{48.83}    & \textbf{82.33 }   & \textbf{73.04}    & \textbf{87.66}   \\  \bottomrule
            \end{tabular}
        }
        \label{tab:unseen}
    \end{table}

    \begin{table}[]
        \caption{Linear-Accuracy ($\%$) with respect to different combinations of the proposed IM and UM matrices in the ATS module. 
        The incorporation of two matrices improves the performance greatly. $*$ denotes that the coefficients are learnable instead manually set.}
        \resizebox{\linewidth}{!}{
            \begin{tabular}{ccc|ccccc}
            \toprule
                $\alpha$   & $\beta$    & $\gamma$    & PKU I  & PKU II  & NTU 60 & NTU 120 & UESTC  \\ 
            \midrule
                1          & 0          & 0           & 77.77  & 48.89   & 81.08  & 74.77   & 80.91 \\
                1          & 1          & 0           & 79.19  & 50.76   & 82.41  & 78.35   & 84.32 \\
                1          & 0          & 1           & 81.23  & 53.40   & 83.38  & 78.23   & 89.68 \\
                1          & 1          & 1           & 82.31  & 55.93   & 83.50  & 78.13   & 91.20 \\
            \cdashline{1-8}
                \rowcolor{gray!15} $*$         & $*$        & $*$           & \textbf{82.79}  & \textbf{57.02}   & \textbf{84.77}  & \textbf{79.41}   & \textbf{91.33} \\
            \bottomrule
            \end{tabular}
        }
        \label{tab:ats}
        \vspace{-8pt}
    \end{table}

    \subsection{General Results}

    \textbf{Comparison with FL methods.} 
    In ~\cref{tab:general_results}, we replace the aggregation strategy FedAVG~\cite{fedavg} in Vanilla FSAR with common FL optimization algorithms: FedProx~\cite{fedprox}, FedBN~\cite{fedbn}, MOON~\cite{moon}, and FedAGM~\cite{fedagm}. 
    Results show that most of the existing FL algorithms are beneficial for improving the performance of Vanilla FSAR, while the gains from them are limited. 
    The accuracy gains they bring are almost around $2\%$. 
    This is because these algorithms lack the consideration of handling heterogeneous skeleton structures. 
    On the contrary, by revisiting human topology graphs and disentangling specific skeleton joint connection knowledge, our FSAR built by ATS and MKD outperforms varying combinations of FL methods under the same settings. 
    The test accuracy has a clear improvement of $10.97\%$ on the UESTC dataset, which indicates the effectiveness of FSAR under the federated learning paradigms. 

    \textbf{Scalability to advanced backbones.} 
    To thoroughly validate the scalability of FSAR, three advanced backbones are employed: CTR-GCN~\cite{ctrgcn}, MST-GCN~\cite{cz-aaai21-msgc}, and MS-G3D~\cite{msg3d}. 
    ~\cref{tab:general_results} demonstrates that the improvement of FASR is still significant with stronger baselines, indicating its potential and extensibility under different settings. 

    \textbf{Generalization to unseen datasets.} 
    ~\cref{tab:unseen} demonstrates the superiority of FSAR when directly testing the globally generalized model on the unseen dataset, compared with Vanilla FSAR, which neglects to deal with heterogeneous human typologies. 
    Results indicate that FSAR is capable of learning a generalized global model under privacy protections, which can serve as a strong feature extractor for action representation on other datasets.

    \begin{figure}[t]
        \centering
        \includegraphics[width=1.0\linewidth]{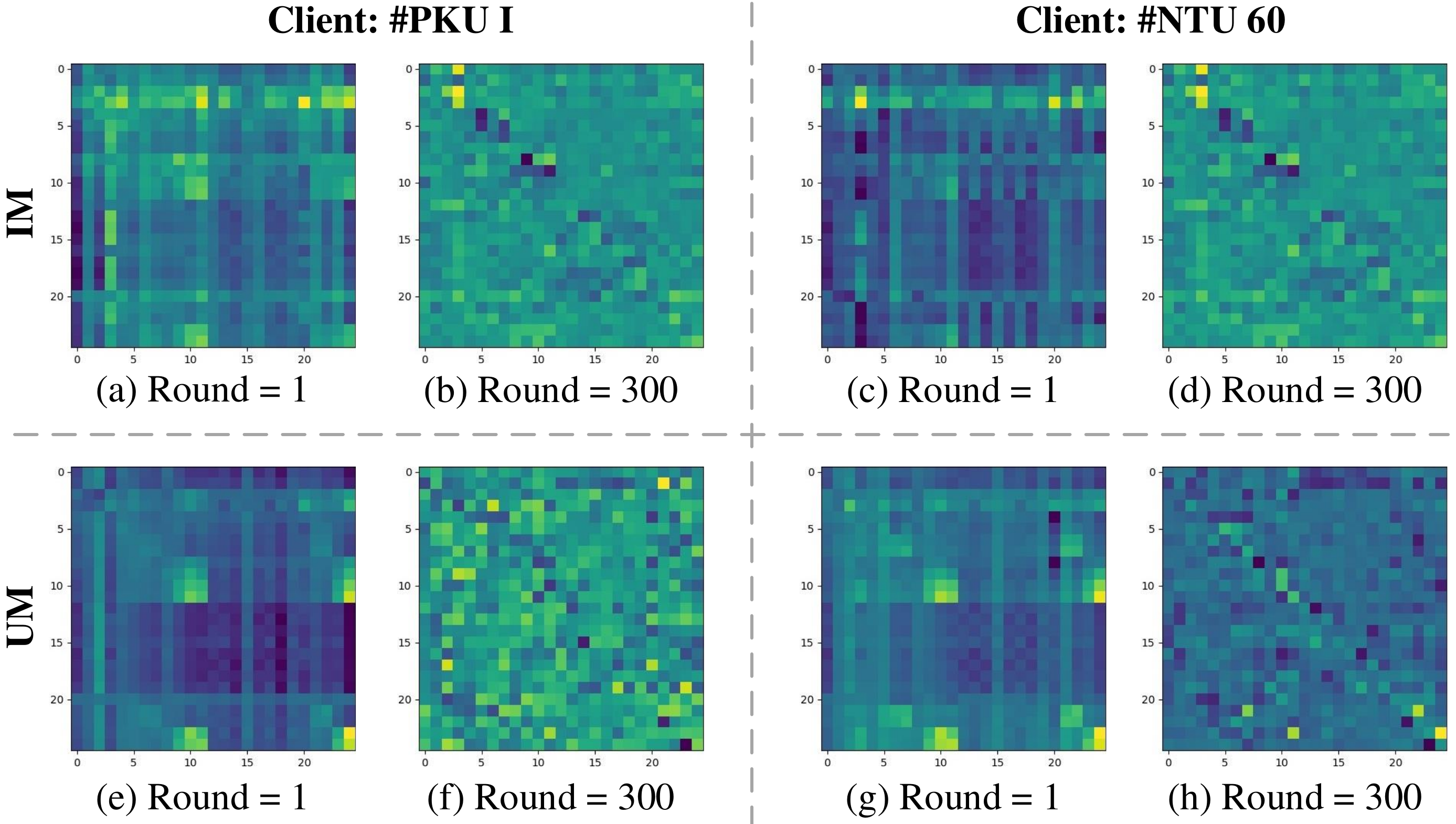}
        \caption{Visualization of the adjacent matrices IM and UM in the ATS module. 
        There exist considerable variations in the similarity of the UM between clients (PKU I, NTU 60) at different training rounds (round 1, round 300), while small for that of the IM.}
        \label{fig:abc}
        \vspace{-8pt}
    \end{figure}

    \subsection{Ablation Studies} 

    \textbf{Effectiveness of different components.} 
    As shown in~\cref{tab:general_results}, we investigate the effects of each module in our model via ablation studies. 
    Compared with FSAR w/o ATS, FSAR w/o MKD can obtain more gains on average, $8.42\%$ on NTU 60, $8.57\%$ on NTU 120, and $9.42\%$ on UESTC, which means that ATS brings slightly more benefit than MKD. 
    The test accuracy curves between different methods in ~\cref{fig:unstable} and ~\cref{fig:test_accuracy} also illustrate that the accession of ATS and MKD largely alleviates the problem of convergence of the training process, since the server aggregates human topology structures from different datasets in an acquirable manner progressively. 
    The above significant enhancements in both numerical comparison and testing curves indicate the significance of mining heterogeneous skeleton topology and decreasing the client-server divergence. 
   
    \textbf{Performance of ATS.} 
    We analyze the effectiveness of IM and UM both qualitatively and quantitatively. 
    From ~\cref{tab:ats}, we can conclude that the UM alone does not gain much for the model, while combined with IM, it further boosts the performance and increases stability. 
    The accession of UM and IM improves $4.54\%$ and $7.04\%$ on PKU I and PKU II datasets, respectively. 
    When changing the factors in~\cref{eq:gcn3} from manually set to learnable, the model can further boost the performance. 
    ~\cref{tab:ats_learnable} reveals that smaller datasets PKU II is equipped with larger factors for IM ($\beta=1.01$), contrast with larger dataset NTU 120 ($\beta=0.67$), which avoids training fluctuations caused by heterogeneity from large datasets and thus achieving more gains ($48.89\%$ to $57.02\%$). 
    We can also find that the $\gamma$ on PKU II and the $\beta$ on NTU 120 vary a lot during the training ($\Delta_{\gamma}=-0.27$ and $\Delta_{\beta}=-0.33$). 
    It can be inferred that $\gamma$ is beneficial for the small-size dataset by enhancing the personalized matrix, and $\beta$ serves the same via weakening the generalized matrix from the large-size dataset.
    Furthermore, ~\cref{fig:abc} demonstrates that the similarity of UM between different clients varies a lot before training (e, g) and after training (f, h), while IM has slight variation. 
    This indicates that IM better mines shared topology graph structures while UM retains personalized knowledge locally.
   
    \begin{table}[]
        \caption{Variation of coefficients ($\Delta$) of the matrices in the ATS module, when they are set learnable and dataset-specific ($\alpha=\beta=\gamma=1$ are set as initialization). 
        The variations of $\gamma$ and $\beta$ are large in PKU II and NTU 120, respectively.}
        \resizebox{\linewidth}{!}{
            \begin{tabular}{c|ccccc|c}
            \toprule
                Coefficient     & PKU I  & PKU II  & NTU 60 & NTU 120 & UESTC & Max/Min $\Delta$\\ 
            \midrule
                $\alpha$   & 1.13   & \textcolor{red}{0.77}    & 1.10   & \textcolor{blue}{0.91}    & 0.85  & (-0.23/-0.09) \\
                $\beta$    & 0.95   & \textcolor{blue}{1.01}   & 0.90   & \textcolor{red}{0.67}     & 0.86  & (-0.33/+0.01) \\
                $\gamma$   & 0.86   & \textcolor{red}{1.27}    & 0.85   & \textcolor{blue}{1.07}    & 1.11  & (-0.27/+0.07) \\
            \bottomrule
            \end{tabular}
        }
        \label{tab:ats_learnable}
        \vspace{-3pt}
    \end{table}

    \begin{figure}[t]
      \centering
        \subfigure[PKU I + UESTC]{\includegraphics[width=4cm]{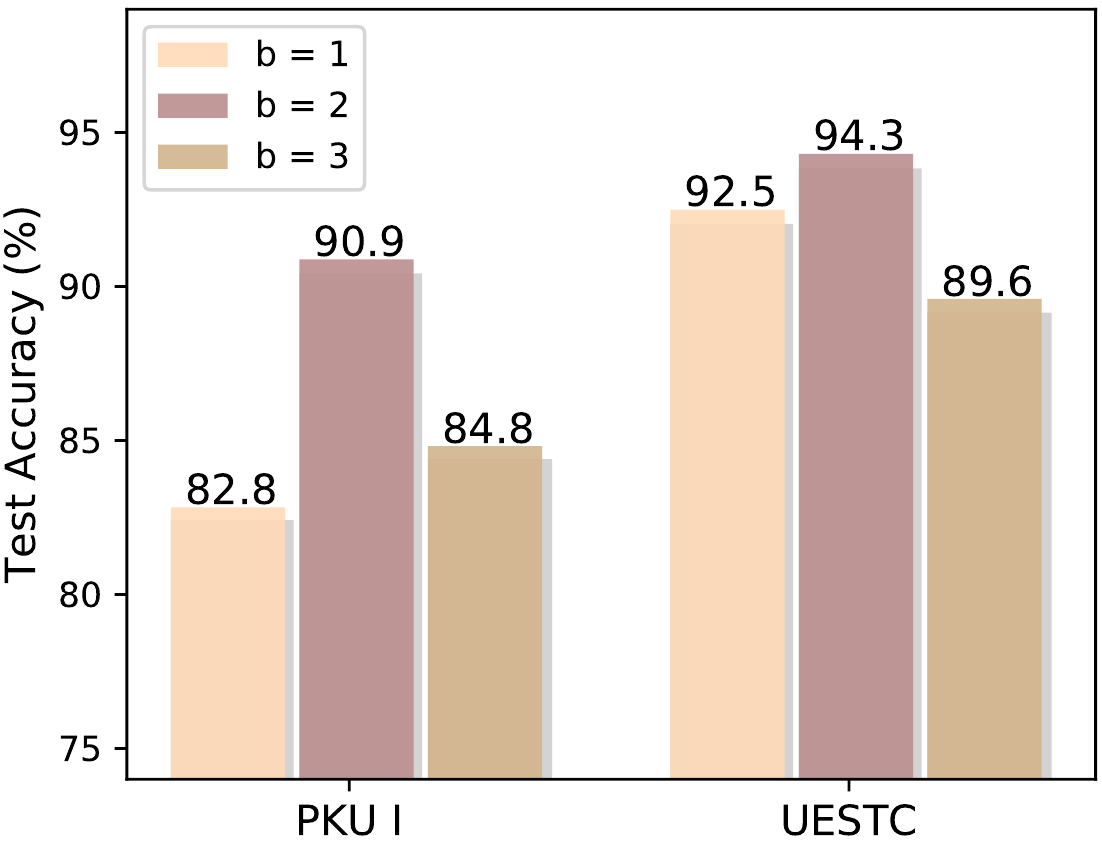}}
        \subfigure[PKU I + NTU 60]{\includegraphics[width=4cm]{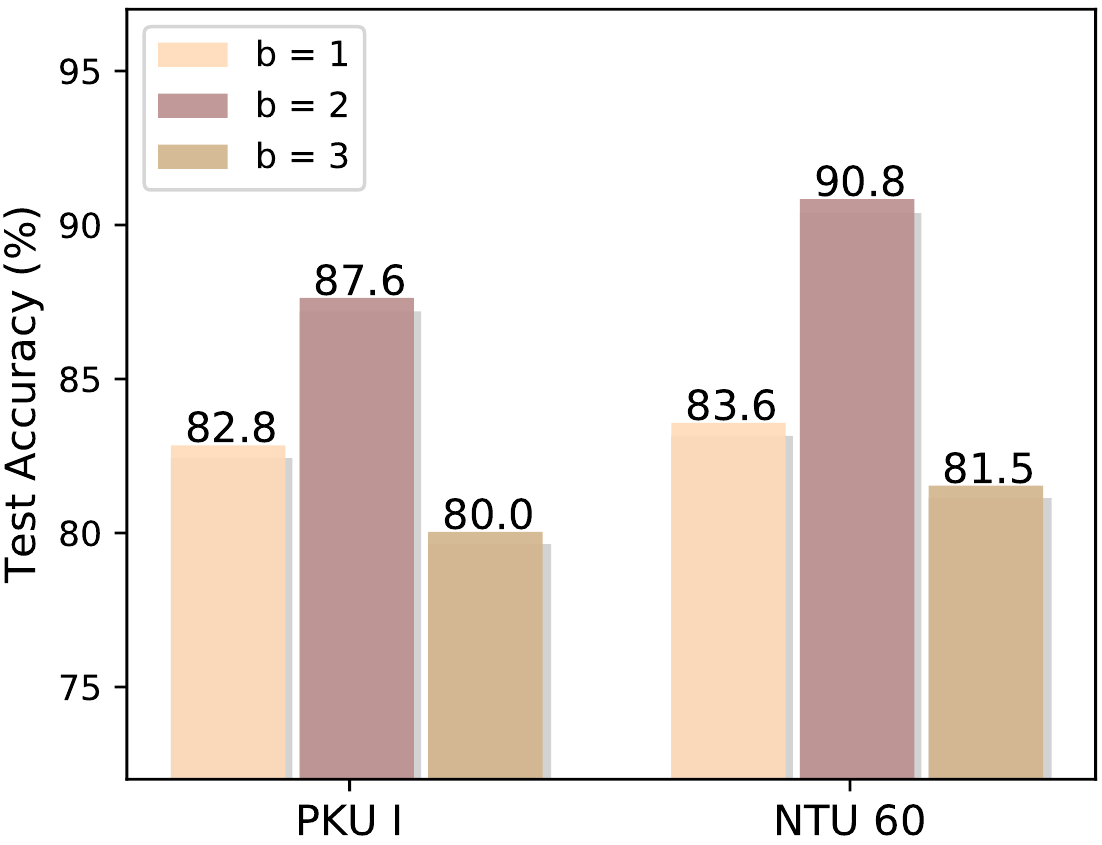}}
      \caption{Ablation studies of different grains of feature from global model considered as previous knowledge to distill clients. Both federated settings achieve the best performance when $b=2$.}
      \label{fig:zft}
      \vspace{-8pt}
    \end{figure}

    \textbf{Performance of MKD.} ~\cref{fig:zft} reports how the number of blocks we align impacts the performance. 
    No loss of generality, we federate PKU I with UESTC and PKU I with NTU 60 to demonstrate the gains. 
    The results show that aligning features from the first two shallow blocks can boost performance most, which achieves an accuracy of $94.3\%$ on UESTC and $90.8\%$ on NTU 60 for the above two settings. 
    This indicates that using different grains of features from shallow blocks of the global model as the teacher knowledge can lead the local model in the right direction. 
    However, aligning deeper blocks cannot bring further improvements (like $b=3$) since deeper features are client-relevant, and these personalized attributes should be retained locally.

    \begin{figure}[t]
      \centering
        \subfigure[Vanilla FSAR]{\includegraphics[width=4cm]{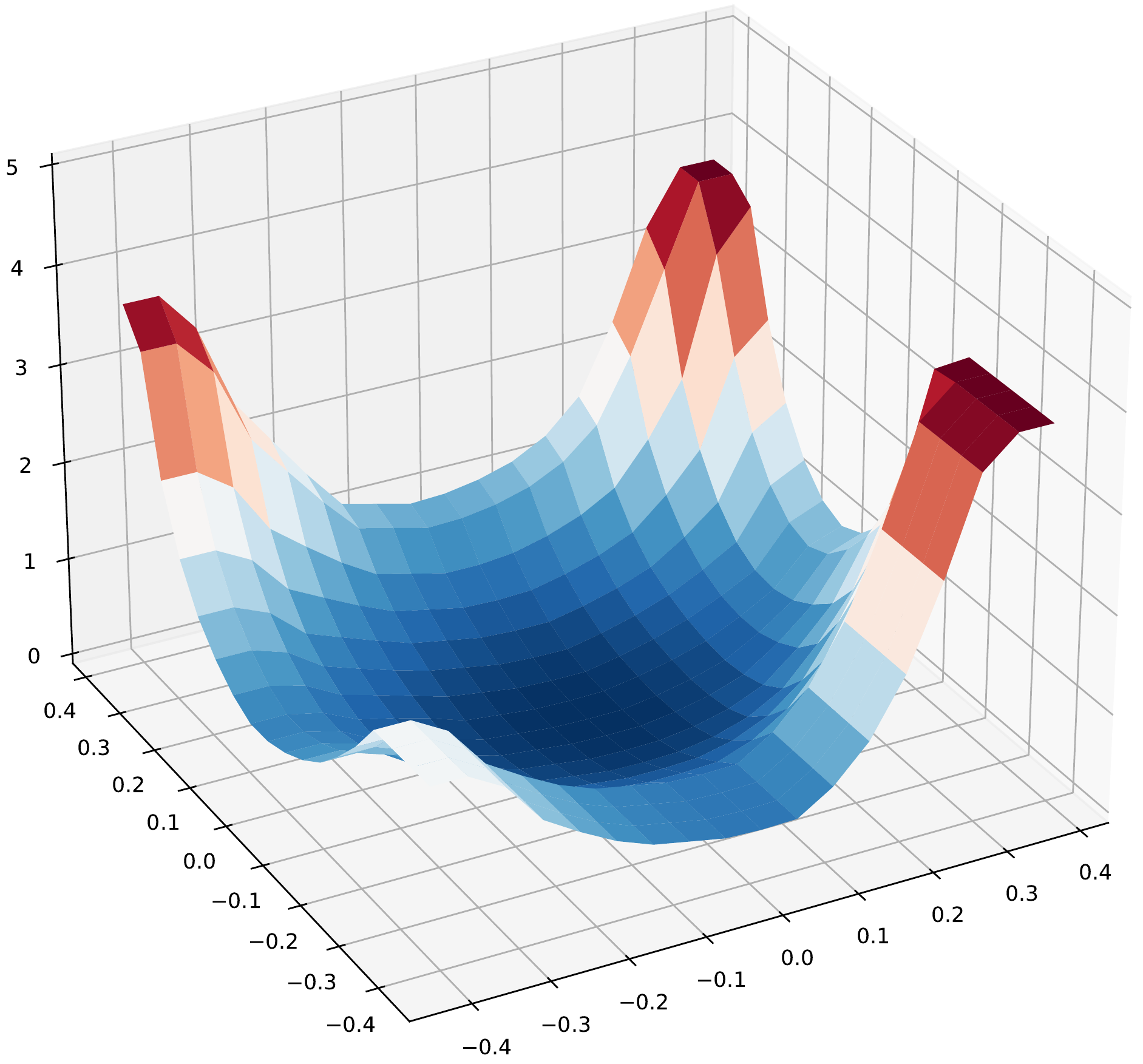}}
        \subfigure[FSAR (Ours)]{\includegraphics[width=4cm]{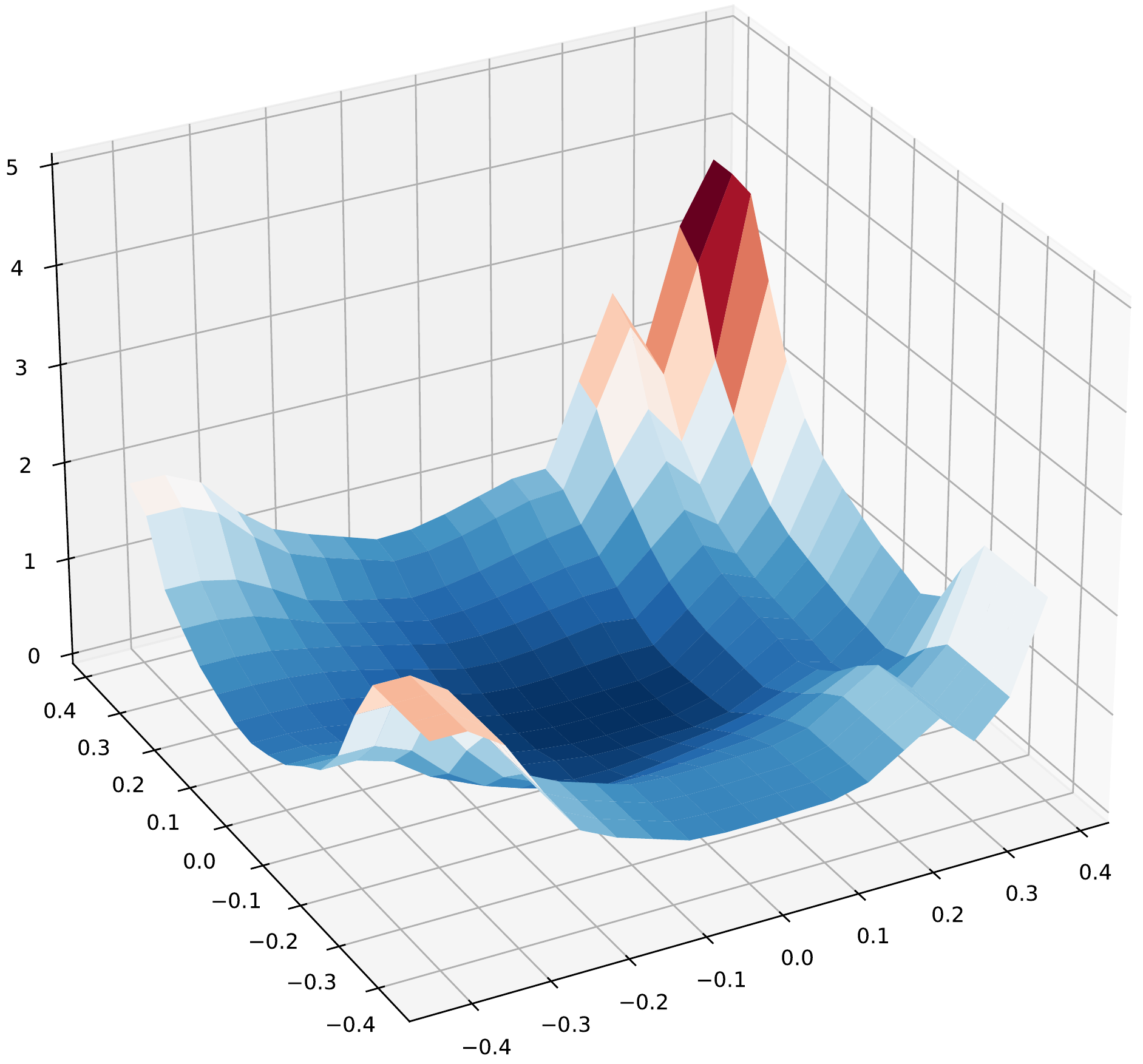}}
      \caption{Visualization of the loss landscape. Compared with Vanilla FSAR, FSAR has a much smoother surface, explaining its superiority in convergence and training stability.}
      \label{fig:losslandscape}
      \vspace{-8pt}
    \end{figure}

    \subsection{Qualitative Analysis}
    To analyze why FSAR can achieve better training stability and convergence than Vanilla FSAR, we further plot the loss landscape~\cite{loss-landscape} of these two models in~\cref{fig:losslandscape}. 
    The figure demonstrates that FSAR has a much smoother loss landscape than Vanilla FSAR, which indicates that the proposed heterogeneous skeleton topology mining and shallow motion features distilling mechanisms can help the individual clients learn more information about others during the client-server communications. This alleviates the negative effect caused by client drift, which further helps the model to find the optimal faster and gain better convergence.

\section{Conclusion}
\label{sec:con}
    This paper takes the lead in introducing FL into skeleton-based action recognition task and present a new paradigm FSAR. 
    We first investigate and discover that the heterogeneous human topology graph structure is the crucial factor hindering training stability. 
    To this end, we introduce an Adaptive Topology Structure module to extract the intrinsic structure of skeleton topology which alleviates unstable training. 
    Furthermore, the Multi-grain Knowledge Distillation mechanism is developed to bridge the divergence between the local clients and the central server. 
    Our proposed benchmark and method provide practical solutions for FL in skeleton-based action recognition and pave the way for future research in this area, which positively contributes to social privacy protection.
    \textbf{Limitations.} 
    There is still room of improvement to adaptively select the depth of feature in MKD for distillation instead of manually set.

{\small
\bibliographystyle{ieee_fullname}
\bibliography{egbib}

\begin{thebibliography}{10}\itemsep=-1pt

\bibitem{ctrgcn}
Yuxin Chen, Ziqi Zhang, Chunfeng Yuan, Bing Li, Ying Deng, and Weiming Hu.
\newblock Channel-wise topology refinement graph convolution for skeleton-based
  action recognition.
\newblock In {\em Proceedings of the IEEE International Conference on Computer
  Vision (ICCV)}, pages 13359--13368, 2021.

\bibitem{cz-aaai21-msgc}
Zhan Chen, Sicheng Li, Bing Yang, Qinghan Li, and Hong Liu.
\newblock Multi-scale spatial temporal graph convolutional network for
  skeleton-based action recognition.
\newblock In {\em Proceedings of the AAAI Conference on Artificial Intelligence
  (AAAI)}, pages 1113--1122, 2021.

\bibitem{gcn-ske-ar}
Ke Cheng, Yifan Zhang, Xiangyu He, Weihan Chen, Jian Cheng, and Hanqing Lu.
\newblock Skeleton-based action recognition with shift graph convolutional
  network.
\newblock In {\em Proceedings of the IEEE Conference on Computer Vision and
  Pattern Recognition (CVPR)}, pages 183--192, 2020.

\bibitem{cnn}
Guilhem Cheron, Ivan Laptev, and Cordelia Schmid.
\newblock P-{CNN}: Pose-based {CNN} features for action recognition.
\newblock In {\em Proceedings of the IEEE International Conference on Computer
  Vision (ICCV)}, pages 3218--3226, 2015.

\bibitem{spact}
Ishan~Rajendrakumar Dave, Chen Chen, and Mubarak Shah.
\newblock S{PA}ct: Self-supervised privacy preservation for action recognition.
\newblock In {\em Proceedings of the IEEE Conference on Computer Vision and
  Pattern Recognition (CVPR)}, pages 20164--20173, 2022.

\bibitem{gcn}
Michael Defferrard, Xavier Bresson, and Pierre Vandergheynst.
\newblock Convolutional neural networks on graphs with fast localized spectral
  filtering.
\newblock {\em Advances in Neural Information Processing Systems (NeurIPS)},
  pages 3837--3845, 2016.

\bibitem{ggcn}
Xiang Gao, Wei Hu, Jiaxiang Tang, Pan Pan, Jiaying Liu, and Zongming Guo.
\newblock Generalized graph convolutional networks for skeleton-based action
  recognition.
\newblock {\em arXiv preprint arXiv:1811.12013}, 2018.

\bibitem{fed-medical-rl}
Pengfei Guo, Dong Yang, Ali Hatamizadeh, An Xu, Ziyue Xu, Wenqi Li, Can Zhao,
  Daguang Xu, Stephanie Harmon, Evrim Turkbey, et~al.
\newblock Auto-{F}ed{RL}: Federated hyperparameter optimization for
  multi-institutional medical image segmentation.
\newblock {\em arXiv preprint arXiv:2203.06338}, 2022.

\bibitem{manual1}
Jian-Fang Hu, Wei-Shi Zheng, Jianhuang Lai, and Jianguo Zhang.
\newblock Jointly learning heterogeneous features for {RGB-D} activity
  recognition.
\newblock In {\em Proceedings of the IEEE Conference on Computer Vision and
  Pattern Recognition (CVPR)}, pages 5344--5352, 2015.

\bibitem{har2}
Shuiwang Ji, Wei Xu, Ming Yang, and Kai Yu.
\newblock 3{D} convolutional neural networks for human action recognition.
\newblock {\em IEEE Transactions on Pattern Analysis and Machine Intelligence
  (TPAMI)}, 35(1):221--231, 2012.

\bibitem{uestc}
Yanli Ji, Feixiang Xu, Yang Yang, Fumin Shen, Heng~Tao Shen, and Wei-Shi Zheng.
\newblock A large-scale {RGB-D} database for arbitrary-view human action
  recognition.
\newblock In {\em Proceedings of the ACM International Conference on Multimedia
  (ACM MM)}, pages 1510--1518, 2018.

\bibitem{fl-kairouz}
Peter Kairouz, H~Brendan McMahan, Brendan Avent, Aurelien Bellet, Mehdi Bennis,
  Arjun~Nitin Bhagoji, Kallista Bonawitz, Zachary Charles, Graham Cormode,
  Rachel Cummings, et~al.
\newblock Advances and open problems in federated learning.
\newblock {\em Foundations and Trends in Machine Learning (FTML)},
  14(1--2):1--210, 2021.

\bibitem{client-drift}
Sai~Praneeth Karimireddy, Satyen Kale, Mehryar Mohri, Sashank Reddi, Sebastian
  Stich, and Ananda~Theertha Suresh.
\newblock S{CAFFOLD}: Stochastic controlled averaging for federated learning.
\newblock In {\em Proceedings of the International Conference on Machine
  Learning (ICML)}, pages 5132--5143, 2020.

\bibitem{fedagm}
Geeho Kim, Jinkyu Kim, and Bohyung Han.
\newblock Communication-efficient federated learning with acceleration of
  global momentum.
\newblock {\em arXiv preprint arXiv:2201.03172}, 2022.

\bibitem{har1}
Yu Kong and Yun Fu.
\newblock Human action recognition and prediction: A survey.
\newblock {\em International Journal of Computer Vision (IJCV)},
  130(5):1366--1401, 2022.

\bibitem{cka}
Simon Kornblith, Mohammad Norouzi, Honglak Lee, and Geoffrey Hinton.
\newblock Similarity of neural network representations revisited.
\newblock In {\em Proceedings of the International Conference on Machine
  Learning (ICML)}, pages 3519--3529, 2019.

\bibitem{rnn}
Guy Lev, Gil Sadeh, Benjamin Klein, and Lior Wolf.
\newblock {RNN} fisher vectors for action recognition and image annotation.
\newblock In {\em Proceedings of the European Conference on Computer Vision
  (ECCV)}, pages 833--850, 2016.

\bibitem{loss-landscape}
Hao Li, Zheng Xu, Gavin Taylor, Christoph Studer, and Tom Goldstein.
\newblock Visualizing the loss landscape of neural nets.
\newblock {\em Advances in Neural Information Processing Systems (NeurIPS)},
  pages 6391--6401, 2018.

\bibitem{gcn-ske-ar2}
Maosen Li, Siheng Chen, Xu Chen, Ya Zhang, Yanfeng Wang, and Qi Tian.
\newblock Actional-structural graph convolutional networks for skeleton-based
  action recognition.
\newblock In {\em Proceedings of the IEEE Conference on Computer Vision and
  Pattern Recognition (CVPR)}, pages 3595--3603, 2019.

\bibitem{moon}
Qinbin Li, Bingsheng He, and Dawn Song.
\newblock Model-contrastive federated learning.
\newblock In {\em Proceedings of the IEEE Conference on Computer Vision and
  Pattern Recognition (CVPR)}, pages 10713--10722, 2021.

\bibitem{fedprox}
Tian Li, Anit~Kumar Sahu, Manzil Zaheer, Maziar Sanjabi, Ameet Talwalkar, and
  Virginia Smith.
\newblock Federated optimization in heterogeneous networks.
\newblock {\em Proceedings of Machine Learning and Systems (MLSys)}, pages
  429--450, 2020.

\bibitem{fedbn}
Xiaoxiao Li, Meirui Jiang, Xiaofei Zhang, Michael Kamp, and Qi Dou.
\newblock Fed{BN}: Federated learning on non-{IID} features via local batch
  normalization.
\newblock {\em arXiv preprint arXiv:2102.07623}, 2021.

\bibitem{person-identify-2}
Rijun Liao, Chunshui Cao, Edel~B Garcia, Shiqi Yu, and Yongzhen Huang.
\newblock Pose-based temporal-spatial network ({PTSN}) for gait recognition
  with carrying and clothing variations.
\newblock In {\em Proceedings of the Chinese Conference on Biometric
  Recognition (CCBR)}, pages 474--483, 2017.

\bibitem{person-identify}
Rijun Liao, Shiqi Yu, Weizhi An, and Yongzhen Huang.
\newblock A model-based gait recognition method with body pose and human prior
  knowledge.
\newblock {\em Pattern Recognition (PR)}, 98:107069, 2020.

\bibitem{pkummd}
Chunhui Liu, Yueyu Hu, Yanghao Li, Sijie Song, and Jiaying Liu.
\newblock {PKU-MMD}: A large scale benchmark for continuous multi-modal human
  action understanding.
\newblock {\em arXiv preprint arXiv:1703.07475}, 2017.

\bibitem{nturgbd120}
Jun Liu, Amir Shahroudy, Mauricio Perez, Gang Wang, Ling-Yu Duan, and Alex~C
  Kot.
\newblock {NTU} {RGB+D} 120: A large-scale benchmark for 3{D} human activity
  understanding.
\newblock {\em IEEE Transactions on Pattern Analysis and Machine Intelligence
  (TPAMI)}, 42(10):2684--2701, 2019.

\bibitem{cnn-liumengyuan}
Mengyuan Liu, Hong Liu, and Chen Chen.
\newblock Enhanced skeleton visualization for view invariant human action
  recognition.
\newblock {\em Pattern Recognition (PR)}, 68:346--362, 2017.

\bibitem{gcn-ske-ar3}
Ziyu Liu, Hongwen Zhang, Zhenghao Chen, Zhiyong Wang, and Wanli Ouyang.
\newblock Disentangling and unifying graph convolutions for skeleton-based
  action recognition.
\newblock In {\em Proceedings of the IEEE Conference on Computer Vision and
  Pattern Recognition (CVPR)}, pages 143--152, 2020.

\bibitem{msg3d}
Ziyu Liu, Hongwen Zhang, Zhenghao Chen, Zhiyong Wang, and Wanli Ouyang.
\newblock Disentangling and unifying graph convolutions for skeleton-based
  action recognition.
\newblock In {\em Proceedings of the IEEE Conference on Computer Vision and
  Pattern Recognition (CVPR)}, pages 143--152, 2020.

\bibitem{fedavg}
Brendan McMahan, Eider Moore, Daniel Ramage, Seth Hampson, and Blaise~Aguera y
  Arcas.
\newblock Communication-efficient learning of deep networks from decentralized
  data.
\newblock In {\em Proceedings of the International Conference on Artificial
  Intelligence and Statistics (AISTATS)}, pages 1273--1282, 2017.

\bibitem{flhargcn}
Riktim Mondal, Debadyuti Mukherjee, Pawan~Kumar Singh, Vikrant Bhateja, and Ram
  Sarkar.
\newblock A new framework for smartphone sensor-based human activity
  recognition using graph neural network.
\newblock {\em IEEE Sensors Journal (JSEN)}, 21(10):11461--11468, 2020.

\bibitem{pytorch}
Adam Paszke, Sam Gross, Francisco Massa, Adam Lerer, James Bradbury, Gregory
  Chanan, Trevor Killeen, Zeming Lin, Natalia Gimelshein, Luca Antiga, et~al.
\newblock Pytorch: An imperative style, high-performance deep learning library.
\newblock {\em Advances in Neural Information Processing Systems (NeurIPS)},
  pages 8024--8035, 2019.

\bibitem{fedvssl}
Yasar Abbas~Ur Rehman, Yan Gao, Jiajun Shen, Pedro Porto~Buarque de Gusmao, and
  Nicholas Lane.
\newblock Federated self-supervised learning for video understanding.
\newblock In {\em Proceedings of the European Conference on Computer Vision
  (ECCV)}, pages 506--522, 2022.

\bibitem{grafehty}
Abhishek Sarkar, Tanmay Sen, and Ashis~Kumar Roy.
\newblock Gra{F}e{HT}y: Graph neural network using federated learning for human
  activity recognition.
\newblock In {\em Proceedings of the IEEE International Conference on Machine
  Learning and Applications (ICMLA)}, pages 1124--1129, 2021.

\bibitem{nturgbd}
Amir Shahroudy, Jun Liu, Tian-Tsong Ng, and Gang Wang.
\newblock {NTU} {RGB+D}: A large scale dataset for 3{D} human activity
  analysis.
\newblock In {\em Proceedings of the IEEE Conference on Computer Vision and
  Pattern Recognition (CVPR)}, pages 1010--1019, 2016.

\bibitem{agcn}
Lei Shi, Yifan Zhang, Jian Cheng, and Hanqing Lu.
\newblock Two-stream adaptive graph convolutional networks for skeleton-based
  action recognition.
\newblock In {\em Proceedings of the IEEE Conference on Computer Vision and
  Pattern Recognition (CVPR)}, pages 12026--12035, 2019.

\bibitem{asgcn}
Lei Shi, Yifan Zhang, Jian Cheng, and Hanqing Lu.
\newblock Skeleton-based action recognition with multi-stream adaptive graph
  convolutional networks.
\newblock {\em IEEE Transactions on Image Processing (TIP)}, 29:9532--9545,
  2020.

\bibitem{dl-ske-st2}
Xiangbo Shu, Liyan Zhang, Guo-Jun Qi, Wei Liu, and Jinhui Tang.
\newblock Spatiotemporal co-attention recurrent neural networks for
  human-skeleton motion prediction.
\newblock {\em IEEE Transactions on Pattern Analysis and Machine Intelligence
  (TPAMI)}, 44(6):3300--3315, 2021.

\bibitem{rnn-sichengyang}
Chenyang Si, Wentao Chen, Wei Wang, Liang Wang, and Tieniu Tan.
\newblock An attention enhanced graph convolutional {LSTM} network for
  skeleton-based action recognition.
\newblock In {\em Proceedings of the IEEE Conference on Computer Vision and
  Pattern Recognition (CVPR)}, pages 1227--1236, 2019.

\bibitem{fed-reid-2}
Shitong Sun, Guile Wu, and Shaogang Gong.
\newblock Decentralised person re-identification with selective knowledge
  aggregation.
\newblock {\em arXiv preprint arXiv:2110.11384}, 2021.

\bibitem{dl-ske-st}
Juanhui Tu, Mengyuan Liu, and Hong Liu.
\newblock Skeleton-based human action recognition using spatial temporal 3{D}
  convolutional neural networks.
\newblock In {\em Proceedings of the IEEE International Conference on
  Multimedia and Expo (ICME)}, pages 1--6, 2018.

\bibitem{manual2}
Raviteja Vemulapalli, Felipe Arrate, and Rama Chellappa.
\newblock Human action recognition by representing 3{D} skeletons as points in
  a lie group.
\newblock In {\em Proceedings of the IEEE Conference on Computer Vision and
  Pattern Recognition (CVPR)}, pages 588--595, 2014.

\bibitem{har3}
Jiang Wang, Zicheng Liu, Ying Wu, and Junsong Yuan.
\newblock Learning actionlet ensemble for 3{D} human action recognition.
\newblock {\em IEEE Transactions on Pattern Analysis and Machine Intelligence
  (TPAMI)}, 36(5):914--927, 2013.

\bibitem{fed-reid}
Guile Wu and Shaogang Gong.
\newblock Decentralised learning from independent multi-domain labels for
  person re-identification.
\newblock In {\em Proceedings of the AAAI Conference on Artificial Intelligence
  (AAAI)}, pages 2898--2906, 2021.

\bibitem{fed-medical}
An Xu, Wenqi Li, Pengfei Guo, Dong Yang, Holger~R Roth, Ali Hatamizadeh, Can
  Zhao, Daguang Xu, Heng Huang, and Ziyue Xu.
\newblock Closing the generalization gap of cross-silo federated medical image
  segmentation.
\newblock In {\em Proceedings of the IEEE Conference on Computer Vision and
  Pattern Recognition (CVPR)}, pages 20866--20875, 2022.

\bibitem{stgcn}
Sijie Yan, Yuanjun Xiong, and Dahua Lin.
\newblock Spatial temporal graph convolutional networks for skeleton-based
  action recognition.
\newblock In {\em Proceedings of the AAAI Conference on Artificial Intelligence
  (AAAI)}, pages 7444--7452, 2018.

\bibitem{noniid}
Yue Zhao, Meng Li, Liangzhen Lai, Naveen Suda, Damon Civin, and Vikas Chandra.
\newblock Federated learning with non-iid data.
\newblock {\em arXiv preprint arXiv:1806.00582}, 2018.

\bibitem{fed-vln}
Kaiwen Zhou and Xin~Eric Wang.
\newblock Fed{VLN}: Privacy-preserving federated vision-and-language
  navigation.
\newblock {\em arXiv preprint arXiv:2203.14936}, 2022.

\bibitem{fedema}
Weiming Zhuang, Yonggang Wen, and Shuai Zhang.
\newblock Divergence-aware federated self-supervised learning.
\newblock {\em arXiv preprint arXiv:2204.04385}, 2022.

\end{thebibliography}
}

\end{document}